\documentclass[lettersize,journal]{IEEEtran}
\newcommand{\arrowlocation}{0.5}

\usepackage{amsmath,amsfonts}
\usepackage{algorithmic}
\usepackage{algorithm}
\usepackage{array}
\usepackage[caption=false,font=footnotesize,labelfont=rm,textfont=rm]{subfig}
\usepackage{textcomp}
\usepackage{stfloats}
\usepackage{url}
\usepackage{verbatim}
\usepackage{graphicx}
\usepackage{cite}
\usepackage{color}
\usepackage{multirow}
\usepackage{booktabs}
\usepackage{diagbox}

\usepackage{tikz}
\usepackage[export]{adjustbox}
\usepackage{dsfont}
\usepackage[T1]{fontenc} 

\begin{document}
\title{Learning to Scale Temperature in Masked Self-Attention for Image Inpainting}
\author{Xiang Zhou, Yi Gong, and Yuan Zeng
\thanks{Xiang Zhou and Yi Gong are with Department of Electronic and Electrical Engineering, Southern University of Science and Technology, Shenzhen, China. E-mail: zhoux2020@mail.sustech.edu.cn; gongy@sustech.edu.cn}
\thanks{Yuan Zeng is with Academy for Advanced Interdisciplinary Studies, Southern University of Science and Technology, Shenzhen, China. E-mail: hi.zengyuan@gmail.com}}

\maketitle
\begin{abstract}
 Recent advances in deep generative adversarial networks (GAN) and self-attention mechanism have led to significant improvements in the challenging task of inpainting large missing regions in an image. These methods integrate self-attention mechanism in neural networks to utilize surrounding neural elements based on their correlation and help the networks capture long-range dependencies. Temperature is a parameter in the Softmax function used in the self-attention, and it enables biasing the distribution of attention scores towards a handful of similar patches. Most existing self-attention mechanisms in image inpainting are convolution-based and set the temperature as a constant, performing patch matching in a limited feature space. In this work, we analyze the artifacts and training problems in previous self-attention mechanisms, and redesign the temperature learning network as well as the self-attention mechanism to address them. We present an image inpainting framework with a multi-head temperature masked self-attention mechanism, which provides stable and efficient temperature learning and uses multiple distant contextual information for high quality image inpainting. In addition to improving image quality of inpainting results, we generalize the proposed model to user-guided image editing by introducing a new sketch generation method. Extensive experiments on various datasets such as Paris StreetView, CelebA-HQ and Places2 clearly demonstrate that our method not only generates more natural inpainting results than previous works both in terms of perception image quality and quantitative metrics, but also enables to help users to generate more flexible results that are related to their sketch guidance.
\end{abstract}

\begin{figure*}[t]
    \captionsetup{justification=centering}
    \null\hfill
    \subfloat[Masked Input]{\includegraphics[width=0.11\linewidth]{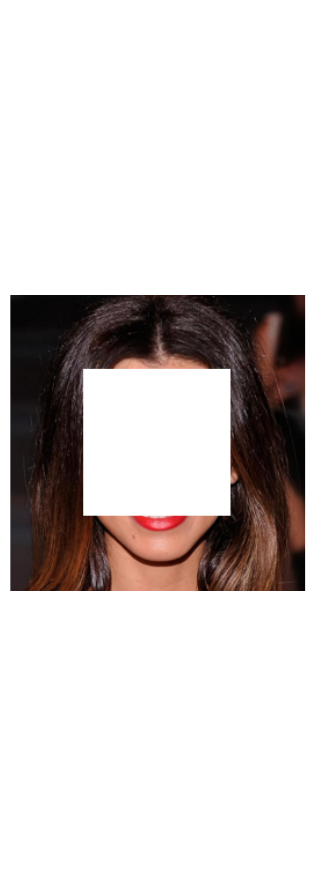}}\hfill
    \subfloat[Coarse Result]{\includegraphics[width=0.11\linewidth]{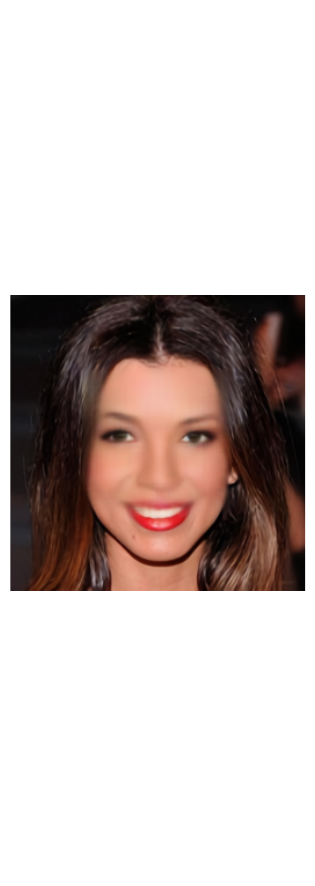}}\hfill
    \subfloat[Temperature Scaling Trajectory]{\includegraphics[width=0.36\linewidth]{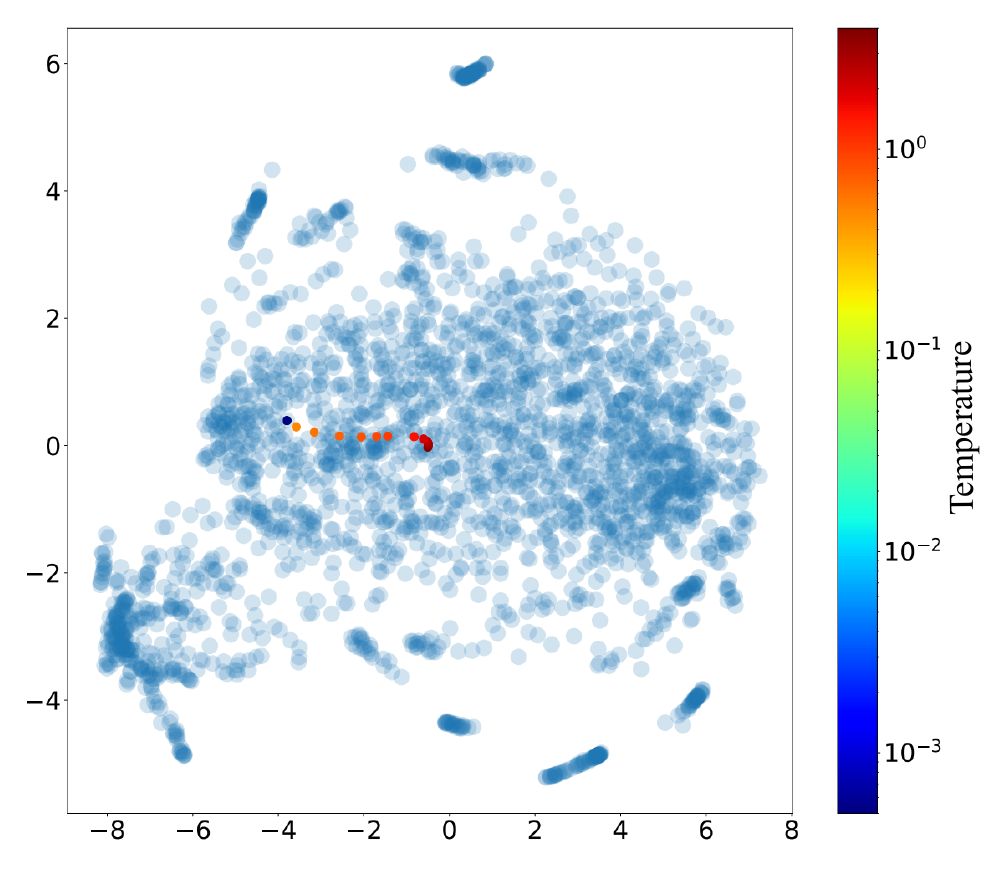}}\hfill
    \subfloat[Temperature Scaling Features]{\includegraphics[width=0.11\linewidth]{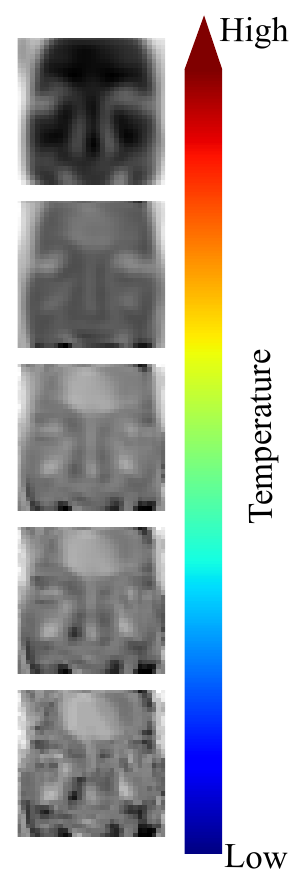}}\hfill
    \subfloat[Final Result]{\includegraphics[width=0.11\linewidth]{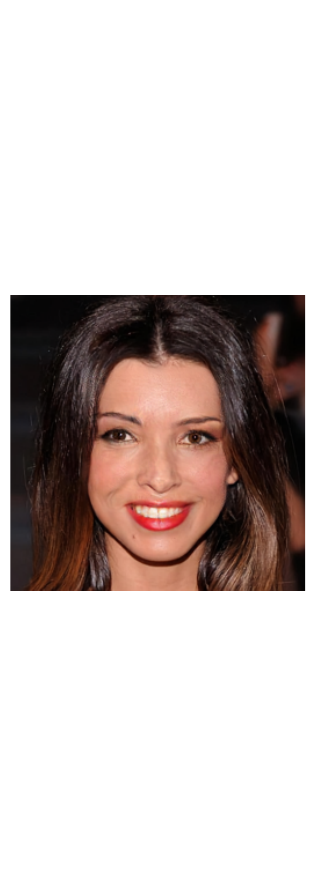}}
    \hfill\null
    \caption{Illustration of temperature scaling in our masked self-attention for image inpainting. In (c), we show the trajectory of a refined query patch in mask region under different temperatures, and the background blue dots represent key patches in known region. The corresponding refined feature maps are shown in (d). As temperature increases, the position of refined patch moves from the most similar key patch position to the center, and the refined feature map becomes smooth. Our module adaptively tunes the temperature to get the final inpainting result.}
    \label{fig:concept}
\end{figure*}
\section{Introduction}
Image inpainting refers to filling holes or masked regions with plausible content coherent with the neighborhood context while making the entire image visually realistic. It allows to remove undesired objects or restore damaged regions in images and serves various computer vision applications, such as object removal, image editing, image denoising and image-based blending. Traditional methods\ \cite{935036, 10.1145/1531326.1531330} infer the pixels of missing regions by propagating from hole boundaries or copying background patches into holes starting from low-resolution to high-resolution. These methods work well for synthesizing plausible stationary textures, but usually fail in non-stationary cases where patterns are unique or non-repetitive. Recent learning-based image inpainting methods treat image inpainting as a conditional generation problem and leverage the advancements in deep generative models, such as Generative Adversarial Networks (GAN)\ \cite{NIPS2014_5ca3e9b1} and Variational Auto-encoders (VAE)\ \cite{9051780}, to infer semantic content. Such methods can learn realistic semantics and textures from a large collection of images and synthesize novel contents in an end-to-end fashion. Deep learning-based image inpainting models can be roughly divided into two categories: one-stage models and two-stage models. One-stage models only include one deep generative network for image completion. Two-stage models consist of a content inference network for coarse image or edge/semantic map completion and a refinement network for high quality appearance.

A key challenge in image inpainting is exploiting visible information globally when holes are large, or the expected contents inside holes have complicated semantic layouts, textures, or depths. 
To address this challenge, more recent two-stage image inpainting methods\ \cite{8578675, zheng2021tfill, zhou2021atma} exploited self-attention to increase the receptive field with desired contextual information in the second stage. Attention mechanisms in neural networks explicitly model the relation between neural elements based on their correlation, helping deep learning introduce long-range dependencies. They share similar principles to non-local processing, and serve as a crucial component in various natural language processing and computer vision tasks. For image inpainting, the self-attention mechanism is very closely linked to the problem of nearest neighbor-based patch matching. It performs non-local patch matching via replacing the filtering of matched patches with a convolution layer, and uses Softmax function to transform similarity scores into probabilities. Temperature is a hyperparameter used in Softmax function for biasing the distribution of weights towards a handful of similar patches. Although significant progress has been made in improving the quality of image inpainting, attention-based image inpainting models commonly set the temperature to a constant, making attention on limited spatial locations in feature space. A possible solution is to learn the temperature and adaptively attend on similar patches at multiple distant spatial locations. In\ \cite{zhou2021atma}, we proposed an attention mechanism, called adaptive multi-temperature mask-guided attention (ATMA), and showed that ATMA can help GAN model generate higher quality inpainting results. However, ATMA is a convolution-based self-attention mechanism which limits the size of patch and requires iterative computation. Furthermore, its temperature learning process is not stable during training, and the quality of its results including color discrepancy and edge responses surrounding holes, needs further improvement. This work aims to propose a new attention mechanism, improve the training stability of the temperature learning and further reduce visual artifacts.

Similar to contextual attention in\ \cite{8578675}, ATMA performed patch matching as convolution-based attention. It used key patches as convolution kernels that can be properly normalized for cosine similarity calculation, while query patches remained not normalized due to limitation of convolution-based attention calculation. For low constant temperature, such mechanisms work well on patch matching. While for learning temperature, query patches should be normalized to prevent length of queries affect on temperature learning. In contrast to doing convolution-based self-attention, this work proposes a masked self-attention mechanism that does dot-product attention with queries, keys and values. By normalizing queries and keys, the similarity comparison can be implemented with simple matrix multiplications in parallel. The proposed attention module enables the use of patches of any size, and provides efficient temperature learning.

For the temperature training stability and visual artifacts in ATMA\ \cite{zhou2021atma}, we identify the causes for these issues and describe changes in architecture that eliminate them. First, we investigate the origin of the temperature training stability problem. We find that the LeakyReLU activation function used in the last layer of the temperature learning network may produce negative temperature, resulting in the largest weight on the most dissimilar neural patch and unstable training of temperature. In Section\ \ref{tem}, we redesign the activation function, which provides training with stabilized temperature learning network against negative temperatures. These changes in architecture are simple, fast and stable in training, producing high quality inpainting results for free-form holes.

In this work, we carefully intervene in each component of the coarse-to-fine image inpainting framework in previous work\ \cite{zhou2021atma} to alleviate the problem and unlock the potential of the adaptive self-attention for better image inpainting. We propose a novel free-form image inpainting framework with a multi-head temperature masked self-attention mechanism to improve training stability and inpainting quality. The main contributions of this work are summarized as follows:
\begin{itemize}
\item A novel multi-head temperature masked self-attention (MHTMA) mechanism that exploits context information to ensure appearance consistency is introduced and integrated into a coarse-to-fine image inpainting framework. It addresses high quality image inpainting by optimizing the feature space for matching, and it utilizes learnable temperatures to attend on matched patches at multiple distant spatial locations, as shown in Fig. \ref{fig:concept}. In addition, it enables parallel non-local patch similarity comparison and provides efficient temperature learning with normalized queries, keys and values.    
\item We redesign a temperature learning network with a new activation function for stable training. During training, the temperature parameter which controls the weight distribution in the masked self-attention is directly optimized.
\item We conduct an in-depth analysis of the effectiveness of different temperature scaling strategies and network branches on inpainting performance. Experimental results show the proposed method can improve image generation quality by learning to control the confidence of using similar neural patches for patch propagation. 
\item We demonstrate the proposed inpainting framework achieves higher quality free-form inpainting than previous state-of-the-art on challenge inpainting datasets, including CelebA-HQ and Places2. 
\end{itemize}


\section{Related Work}
\subsection{Image Inpainting}
Traditional image inpainting methods include diffusion-based and patch-based methods. Diffusion-based methods use variational algorithms 
to propagate color information from the known regions to the holes. These methods work well for small or narrow missing regions, but they tend to result in over-blurring when the missing regions grow larger. Patch-based methods perform texture synthesis techniques to copy-paste neighboring patches from the background regions into the missing regions\ \cite{10.1145/1531326.1531330, 790383}. Compared to diffusion-based methods, they can fill in large missing regions with stationary texture and result in high quality texture. Criminisi et al.\ \cite{1323101} proposed to optimize patch search using multiple scales and orientations. In \cite{10.1145/1531326.1531330}, an approximate nearest neighbor algorithm, called PatchMatch, was proposed and has shown significant practical values for image inpainting. To further reduce computational complexity of patch matching and improve search speed, a coherency sensitive hashing-based image inpainting approach was presented in\ \cite{6126421}. Although patch-based methods can produce convincing continuations of the background, they cannot generate semantic or novel content. 

Recently, many learning-based single-stage image inpainting methods have been proposed to use deep learning or GAN strategy to fill in missing regions with semantic content. Context encoders\ \cite{7780647} firstly handled $64\times{64}$ sized holes by training deep neural networks with both pixel-wise reconstruction loss and generative adversarial loss. Iizuka et al.\ \cite{10.1145/3072959.3073659} proposed local and global discriminators to improve local texture and overall image layout. Zheng et al.\ \cite{8953195} proposed a variational auto-encoders\ \cite{9051780} (VAE)-based pluralistic image inpainting approach to generate diverse image inpainting results. To better handle irregular holes, a partial convolution was introduced for image inpainting in\ \cite{10.1007/978-3-030-01252-6_6}, where the convolution was masked and re-normalized to use valid pixels only. Suvorov et al.\ \cite{9707077} proposed a resolution-robust large mask inpainting model with fast Fourier convolutions. Wang et al. \cite{9318551} presented a dynamic selection network to avoid the interference of the invalid information from holes and utilize valid information adaptively for image inpainting.

Our work is more closely related to two-stage GAN-based models that infer a coarse image in the first stage and afterwards refine the coarse image with visually pleasant textures in the second stage. Yang et al.\ \cite{8099917} proposed to improve the models in\ \cite{7780647} using multi-scale neural patch synthesis in the second stage. EdgeConnect model\ \cite{9022543} predicted salient edges and afterwards generated completion results guided by the prior edges. Yu et al.\ \cite{8578675} introduced contextual attention that enables trainable patch matching in the second stage to produce higher quality inpainting results, and further improved the model using gated convolution and a patch-based GAN loss for free-form image inpainting in\ \cite{Yu_2019_ICCV}. Zeng et al.\ \cite{ZENG2021108036} performed conventional patch matching in the second stage to copy-past high-frequency missing information from training exemplars and generated diverse high quality outputs. Zheng et al.\ \cite{zheng2021tfill} proposed a two-stage image inpainting model that uses a transformer-based architecture for coarse image inpainting in the first stage and an attention-aware layer for improving appearance consistency between visible and generated regions in the second stage. In\ \cite{9711307}, a transformer-based network was designed to recover pluralistic coherent structures together with coarse textures, while the latter convolutional neural network (CNN) was conducted to enhance the local texture details of coarse priors. Xu et al. \cite{9597484} proposed a Texture Memory-Augmented deep image inpainting framework, where texture generation is guided by a texture memory of patch samples extracted from unmasked regions in the second stage. In\ \cite{zhou2021atma}, we introduced multiple self-adaptive temperature parameters to control the scale of the softness of the contextual attention in the second stage. However, the training process of the model in\ \cite{zhou2021atma} is not stable and the inpainting results of the model still need to be improved, such as color consistency and texture details. In this work, we focus on analyzing the effectiveness of temperature scaling on image inpainting and generating more realistic inpainting results with an improved temperature learning method.

\subsection{Self-Attention}
Self-attention mechanism has been successfully used in machine translation 
, increasing the modeling capacity of deep neural networks by concentrating on crucial features and suppressing unimportant ones. Recently, many studies have explored the effectiveness of self-attention on deep learning-based image generation, proving that distant relationship modeling via attention mechanisms enables learning high-dimensional and complex image distribution to generate more realistic outputs. Zhang et al.\ \cite{pmlr-v97-zhang19d} proposed a self-attention module to reconstruct each feature point using the weighted sum of all feature points. 
Yu et al.\ \cite{8578675} proposed a contextual attention mechanism using a convolution-based self-attention mechanism to attend on related features at distant spatial locations. Liu et al.\ \cite{9009473} proposed a coherent semantic attention layer to construct the correlation between neural features of masked regions. Zheng et al. presented a self-attention layer that exploits short-long term context information to improve appearance consistency in\ \cite{8953195} and modified the attention layer to handle the attention to visible regions separately from masked regions in\ \cite{zheng2021tfill}. In\ \cite{zhou2021atma}, we designed a mask-guided attention layer similar to the attention mechanisms in\ \cite{8578675} and\ \cite{NEURIPS2018_f0e52b27} that enable learnable non-local patch matching, and proposed to learn temperatures to attend on neural features at multiple spatial locations. Unlike existing methods, instead of doing a convolution-based self-attention calculation for trainable patch matching, this work designs a masked self-attention to match patches via matrix multiplication of normalized query and key patches, which provides an efficient way to learn the temperature.


\subsection{Temperature Scaling}
Given a vector $\mathbf{x}=\left(x_{1}, \cdots, x_{N}\right)\in{\mathbb{R}^{N}}$, it can be transformed into a probability vector with the Softmax function as 
\begin{equation}
\mathrm{Softmax}(\mathbf{x}/t)_{i}=\frac{e^{x_{i}/t}}{\sum_{n=1}^{N}e^{x_{n}/t}},
\label{eq:softmax}
\end{equation}
where $\mathrm{Softmax}(\mathbf{x}/t)_{i}\in(0, 1)$, and $\sum_{n=1}^{N}\mathrm{Softmax}(\mathbf{x}/t)_{n}=1$. $t>0$ is a temperature parameter, which is able to adjust the smoothness of the output distribution. The lower the temperature, the sharper or harder the distribution gets, and the larger the temperature, the flatter or softer the distribution will be. Temperature scaling has been shown to be important in supervised learning\ \cite{agarwala2020temperature}, model calibration\ \cite{8953721}, knowledge distillation\ \cite{44873, rajasegaran2020self}, contrastive learning\ \cite{pmlr-v119-chen20j, zhang2021temperature}, image generation\ \cite{8578675, NEURIPS2018_f0e52b27} and neutral language supervision\ \cite{pmlr-v139-radford21a, lin-etal-2018-learning}. Existing temperature scaling approaches can be categorized into three groups: constant temperature adjusting, manually designed temperature tuning and adaptive temperature learning. 

Constant temperature adjusting treats the temperature as a constant and adjusts it before model training. Hinton et al.\ \cite{44873} adopted a high temperature in Softmax function for knowledge distillation. The temperature was applied to show rich similarity structure of a cumbersome model. In\ \cite{NIPS2017_3f5ee243}, a constant temperature was used in a dot-product attention to avoid vanishing gradient.
Yu et al.\ \cite{8578675} proposed a contextual attention for image inpainting, and a low constant temperature was used in contextual attention to weight the similarity between foreground and background patches. Caccia et al.\ \cite{Caccia2020Language} established a quality-diversity evaluation procedure for language generation using temperature tuning over local and global sample metrics.

Manually designed temperature tuning approaches tune the temperature parameter iteratively with pre-designed strategies. Kirkpatrick et al.\ \cite{Kirkpatrick83optimizationby} introduced a temperature in Simulated Annealing to balance the exploration and exploitation trade-off in combinatorial optimization, and used exponentially decreasing annealing strategy to tune the temperature. 
Similar temperature tuning strategy was also used in Boltzmann Machine learning algorithm \cite{ackley1985learning}, and Boltzmann exploration\ \cite{sutton2018reinforcement} in reinforcement learning.

Recently, many approaches have been proposed to adaptively tune temperature during training deep learning models. For example, Lin et al.\ \cite{lin-etal-2018-learning} used a neural network to learn temperature and adaptively control the softness of attention for better neural machine translation performance. Radford et al.\ \cite{pmlr-v139-radford21a} optimized the temperature parameter as a log-parameterized multiplicative scalar for language supervision. Zhang et al.\ \cite{zhang2021temperature} proposed a temperature learning-based approach to generate uncertainty scores for many contrastive methods. Pl\"{o}tz et al.\ \cite{NEURIPS2018_f0e52b27} introduced temperature as a relaxation indicator to the approximation of neural nearest neighbor. In\ \cite{zhou2021atma}, we adopted multiple learnable temperatures in a mask-guided attention to adaptively adjust the softness of the attention, resulting in the attention module attending on matched features with different concentration levels and capturing long-term context for better image completion. In this work, we further analyze the effectiveness of temperature scaling on image inpainting, and improve the temperature learning approach with a multi-head attention and a modified activation function for better robustness during training. 
\begin{figure*}[!ht]
    \centering
    \includegraphics[width=0.85\linewidth]{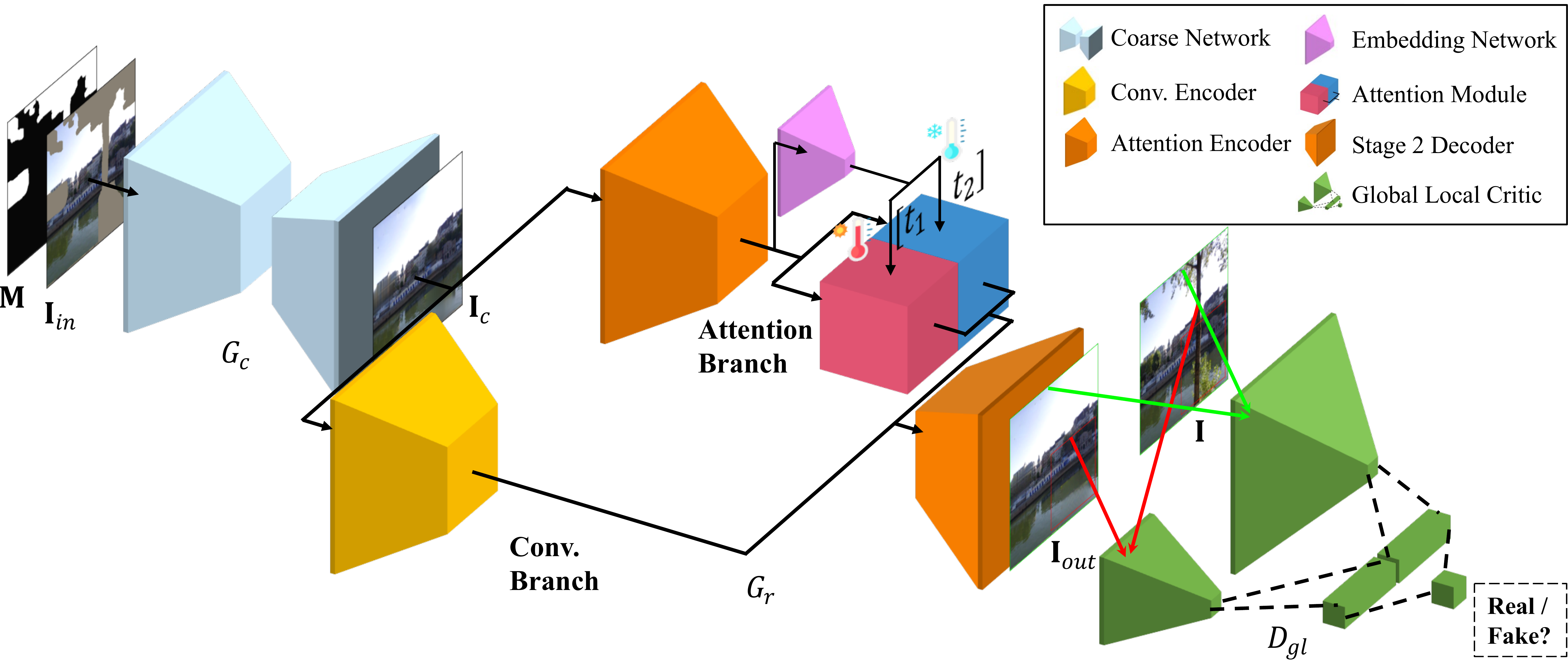}
    \caption{Illustration of our coarse-to-fine image inpainting framework. Given an incomplete image, we use an encoder-decoder network in the first stage to roughly fill in the missing regions. Then, a refinement network is designed in the second stage to capture distant contextual information and aggregate local textual information, which generates final inpainting results with finer texture details and better appearance consistency.}
    \label{fig:architecture}
\end{figure*}
\section{Approach}
This section presents the proposed framework from top to bottom. We first introduce the overview of the proposed framework. Then, we define the loss functions used to train the framework. After that, details of the multi-head temperature masked self-attention and learning-based self-adaptive control of temperature are described.  

\subsection{Coarse-to-fine Inpainting framework}
The overview of the proposed image inpainting framework is illustrated in Fig. \ref{fig:architecture}. Given an input incomplete image $\mathbf{I}_{in}$, which is a degraded version of a complete image $\mathbf{I}_{gt}$. Image inpainting aims at generating visually realistic result $\mathbf{I}_{out}$ from $\mathbf{I}_{in}$ with information of $\mathbf{I}_{gt}$. The proposed inpainting framework is a two-stage coarse-to-fine inpainting architecture. It consists of three parts: a coarse completion network $\mathit{G}_c$ for roughly estimating the missing regions of the input $\mathbf{I}_{in}$ and producing a coarse inpainting result $\mathbf{I}_c$; a refinement network $\mathit{G}_r$ for generating the final output $\mathbf{I}_{out}$ with finer texture and better appearance consistency; a global-local discriminator $\mathit{D}_{gl}$ for calculating adversarial losses. The three parts are sequentially cascaded and trained in an end-to-end manner. 

\textbf{Coarse Completion Network} The coarse completion network $\mathit{G}_c$ is based on a simple encoder-decoder architecture, where gated convolutions \cite{Yu_2019_ICCV} are used at both ends and dilated gated convolutions are used in the middle. Gated convolution learns optimal mask automatically from training data and assigns soft values to each spatial location in deep layers. It performs dynamic feature selection between the mask regions and existing regions for better image inpainting. Dilated gated convolutional layers are used in the mid-layers to compute each output pixel with a much larger input area. The input of the network $\mathit{G}_c$ is an RGB image with free-form holes, and the output $\mathbf{I}_{c}$ is an image with entire content. The encoder extracts neural features of the input $\mathbf{I}_{in}$ and reduces its spatial dimension, and the decoder is adopted to restore the feature dimensions and generate $\mathbf{I}_{c}$.

\textbf{Refinement Network} The intermediate image $\mathbf{I}_{c}$ is then fed into the refinement network $G_r$ for high quality image inpainting. Compared with $\mathbf{I}_{in}$, $\mathbf{I}_{c}$ is more semantically plausible and coherent with known regions, which allows $\mathit{G}_r$ to learn better feature representation for better appearance consistency. $\mathit{G}_r$ consists of two parallel encoders and a single decoder. The first branch encoder tries to capture distant context information using gated convolutions and an improved self-attention module. The second encoder is built with gated and dilated gated convolution layers, and tends to aggregate local texture information. The joint usage of the two encoders provides complementary information from each other. The decoder consists of 5 gated convolution layers and a standard convolution layer. It takes the features extracted from the two encoders as input and generates the final inpainting output $\mathbf{I}_{out}$.  

\textbf{Global-local Discriminator} The output $\mathbf{I}_{out}$ is then discriminated by a global-local  discriminator $D_{gl}$. The discriminator assesses both of the global and local consistency of $\mathbf{I}_{out}$ using two parallel feature extraction branches, including a global branch and a local branch. Specifically, the global branch is performed based on the entire output, guiding the framework to generate a globally coherent inpainting result. It uses 6 convolution layers and a fully-connected layer to extract neural features of the output $\mathbf{I}_{out}$. The local branch consists of 5 convolution layers and a fully-connected layer, and its input is a $128\times{128}$ sized patch that randomly cropped from the output $\mathbf{I}_{out}$. 
The local branch assesses small local areas centered at the completed regions in pursuit of local consistency of the generated patches. After that, the features extracted from the global and local branches are concatenated into a single feature vector and fed into a fully-connected layer. Finally, a sigmoid function is used to distinguish whether the generated output is real or fake.  

\subsection{Loss Functions}
We train our model with a decoupled joint loss function to handle consistency of both high-level semantics and low-level textures. The joint loss includes reconstruction loss $\mathcal{L}^{G}_{r}$, perceptual loss $\mathcal{L}^{G}_p$ and adversarial losses $\mathcal{L}^{D}_{adv}$, $\mathcal{L}^{G}_{adv}$. More details are given as follows.  

\textbf{Reconstruction Loss} We use $l1$ distance to compute the per-pixel reconstruction loss $\mathcal{L}^G_{r}$. The reconstruction loss makes the constraints that the intermediate output $\mathbf{I}_c$ and the final output $\mathbf{I}_{out}$ should approximate the ground-truth $\mathbf{I}_{gt}$, that is,

\begin{equation}
    \mathcal{L}^{G}_{r} = \lambda_1\left\|\mathbf{I}_c-\mathbf{I}_{gt}\right\|_1 + \lambda_2\left\|\mathbf{I}_{out}-\mathbf{I}_{gt}\right\|_1,
\end{equation}
where $\lambda_1$ and $\lambda_2$ are scales.

\textbf{Perceptual Loss} We adopt the perceptual loss \cite{Johnson2016Perceptual} to penalize the perceptual and semantic differences between the intermediate output $\mathbf{I}_{c}$ and the ground-truth $\mathbf{I}_{gt}$, which is given by
\begin{equation}
    \mathcal{L}^{G}_{p} = \left\| \mathcal{V}_{i}(\mathbf{I}_{c}) - \mathcal{V}_{i}(\mathbf{I}_{gt})\right\|_1,
\end{equation}
 where $\mathcal{V}_{i}$ is the activation of the $i$th layer of the VGG-16 network $\mathcal{V}$. The network $\mathcal{V}$ is pre-trained for image classification on ImageNet \cite{5206848} and used as a fixed loss network. In our experiments, we use $\mathrm{ReLU1\_2}$ for feature extraction.
 
\textbf{Adversarial Loss} Motivated by global and local GANs \cite{10.1145/3072959.3073659} and spectral-normalization GANs \cite{DBLP:journals/corr/abs-1802-05957}, we adopt the fast approximation algorithm of spectral normalization described in \cite{DBLP:journals/corr/abs-1802-05957} together with hinge loss \cite{lim2017geometric} to discriminate if the input is real or fake. The adversarial loss functions for discriminator and generator are given by 
\begin{equation}
    \begin{aligned}
        \mathcal{L}_{adv}^D = \mathbb{E}_{\mathbf{I}\sim p_{\mathrm{data}}(\mathbf{I})}&\left[1-D_{gl}(\mathbf{I})\right]_+ \\ &+ \mathbb{E}_{\mathbf{I}_{out}\sim p_{\mathrm{data}}(\mathbf{I}_{out})}\left[1+D_{gl}(\mathbf{I}_{out})\right]_+,
    \end{aligned}
\end{equation}
and
\begin{equation}
    \mathcal{L}_{adv}^G = -\mathbb{E}_{\mathbf{I}_{out}\sim p_{\mathrm{data}}(\mathbf{I}_{out})}\left[D_{gl}(\mathbf{I}_{out})\right],
\end{equation}
where $[\cdot]_{+}$ is a ramp function, i.e.,  $[x]_{+}=\mathrm{max}(0, x)$. $\mathcal{L}_{adv}^D$ and $\mathcal{L}_{adv}^G$ denote the adversarial objective for discriminator and generator, respectively.

\textbf{Full objective} With a weighted sum of the aforementioned losses, our generator is optimized by minimizing the following objective $\mathcal{L}^G$:
\begin{equation}
    \mathcal{L}^G = \mathcal{L}^G_r + \lambda_p\mathcal{L}^G_p + \lambda^G_{adv}\mathcal{L}^{G}_{adv},
\end{equation}
where $\lambda_p$ and $\lambda_{adv}$ are scales. We set $\lambda_1=1.2$, $\lambda_2=1$, $\lambda_p=0.004$ and  $\lambda_{adv}^{G}=0.01$ in our experiments. 

\begin{figure*}[!ht]
    \centering
    \includegraphics[width=0.85\linewidth]{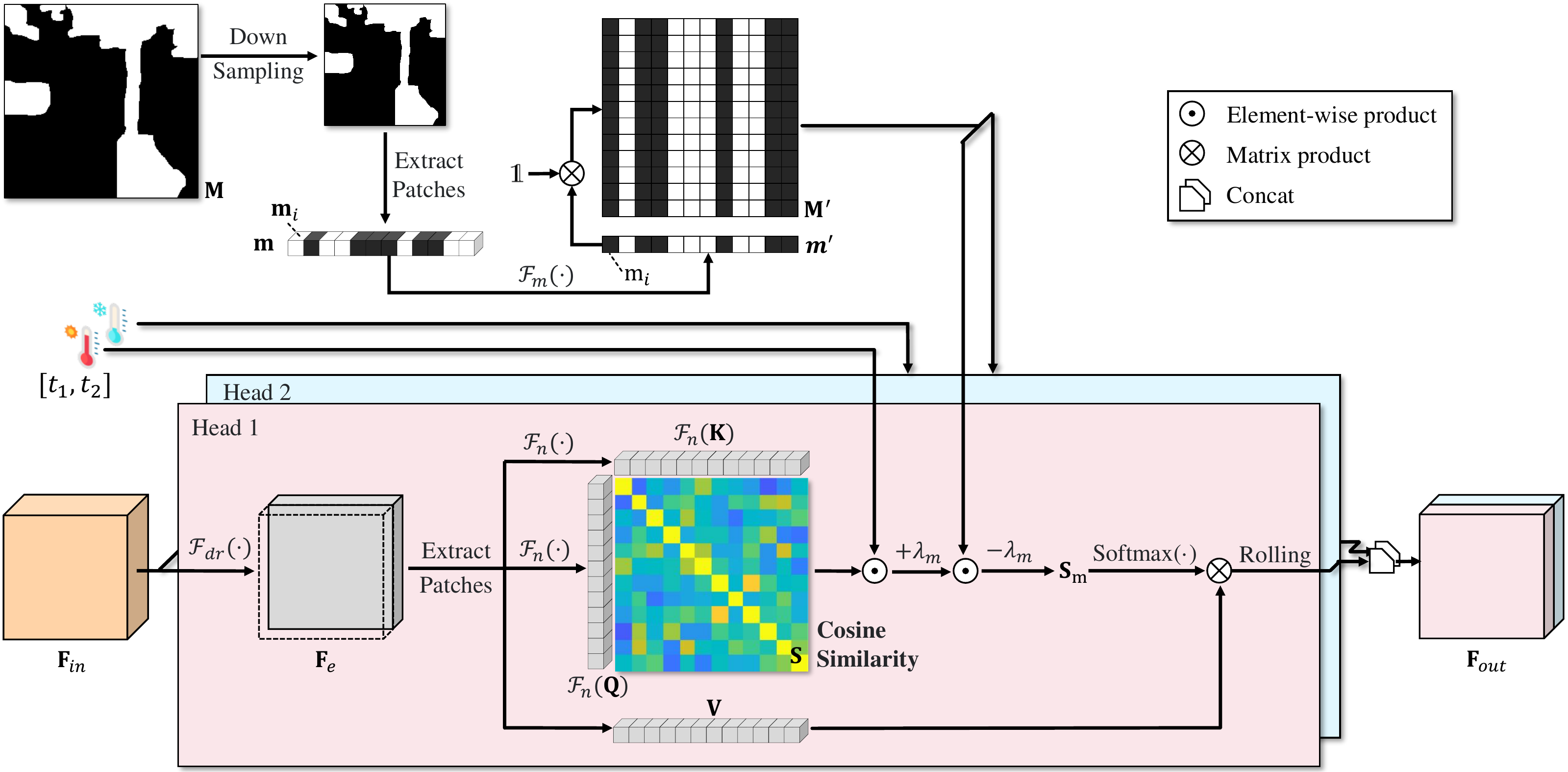}
    \caption{Illustration of our attention module MHTMA.}
    \label{fig:attention}
\end{figure*}

\subsection{Multi-head Temperature Masked Self-attention}
As illustrated in Fig. \ref{fig:attention}, the proposed multi-head temperature masked self-attention (MHTMA) aims at converting the input feature cube $\mathbf{F}_{in}\in\mathbb{R}^{H\times W\times C}$ to a refined feature cube $\mathbf{F}_{out}\in\mathbb{R}^{H\times W\times C}$. 

\textbf{Single Temperature Masked Self-Attention} The input feature cube $\mathbf{F}_{in}$ is first transformed into an embedded feature cube $\mathbf{F}_{e}\in\mathbb{R}^{H\times W\times C^\prime}$ with lower channel dimensions using a channel-wise fully connected network $\mathcal{F}_{dr}(\cdot)$. 

Then, we use unrolling operation \cite{chellapilla:inria-00112631} to extract patches with size of $s\times{s}$ from $\mathbf{F}_{e}$, and get queries $\mathbf{Q}=(\mathbf{q}_1, \dots, \mathbf{q}_{N_q})\in\mathbb{R}^{N_q\times(s \times s \times C^\prime)}$, keys $\mathbf{K}=(\mathbf{k}_1, \dots, \mathbf{k}_{N_k})\in\mathbb{R}^{N_k\times(s \times s \times C^\prime)}$ and values $\mathbf{V}=(\mathbf{v}_1, \dots, \mathbf{v}_{N_k})\in\mathbb{R}^{N_k\times (s \times s \times C^\prime)}$. Note that the number of extracted query patches $N_q$ is different from the number of extracted key patches $N_k$, since zero-padding is performed on feature maps $\mathbf{F}_e$ when extracting query patches to ensure the size of $\mathbf{F}_{out}$ is the same as those of $\mathbf{F}_{in}$. After that, we normalize queries $\mathbf{Q}$ and keys $\mathbf{K}$, and calculate the attention score $\mathbf{S}\in\mathbb{R}^{N_q\times N_k}$ as
\begin{equation}
    \mathbf{S} = \mathcal{F}_n(\mathbf{Q}){\mathcal{F}_n(\mathbf{K}})^T,
\end{equation}
where $\mathcal{F}_n$ denotes normalization.

Let $\mathbf{M}$ denote a binary mask with values 1 for missing regions and 0 for elsewhere. We first down-sample the original mask $\mathbf{M}$ into the size of the input feature cube, which is $H\times{W}$. For each pixel $i$, we extract a patch $\mathbf{m}_{i}$ with size $s\times{s}$ centered on pixel $i$ from the down-sampled mask and transfer the patch $\mathbf{m}_{i}$ into a binary value $m^{\prime}_{i}$, which is given by

\begin{equation}
    m^\prime_{i} = \mathcal{F}_m(\mathrm{\mathbf{m}}_i) = \left\{
        \begin{aligned}
            &1, \quad \mathrm{if}\ \mathrm{sum}(\mathbf{m}_i)=0\\
            &0, \quad \textrm{otherwise}
        \end{aligned}
    \right.
\label{eq:mask}
\end{equation}
After updating mask values for all pixels in the down-sampled mask, we obtain a mask vector $\mathbf{m}^{\prime} \in \mathbb{R}^{N_k\times 1}$. Note that the number of mask patches is the same as the number of keys in $\mathbf{K}$. We construct a new mask $\mathbf{M}^\prime=\mathds{1}(\mathbf{m}^{\prime})^T$ using an all-ones column vector $\mathds{1} \in \mathbb{R}^{N_q\times 1}$, and update the attention score matrix $\mathbf{S}$ as
\begin{equation}
    \label{equation:modify_score}
    \mathbf{S}_m = \mathbf{M}^\prime \odot (\mathbf{S}/t + \lambda_m) - \lambda_m, 
\end{equation}
where $\lambda_m$ is a hyperparameter, $\odot$ denotes element-wise production, $t$ denotes the temperature parameter and $\mathbf{S}_m$ denotes the masked attention score matrix. The masked attention score $\mathbf{S}_m$ is adopted to adjust the attention score to a lower constant level $-\lambda_m$, when keys are extracted from missing regions. After that, Softmax function is employed to transform the masked attention score matrix $\mathbf{S}_m$ into attention coefficients $\mathbf{W}$, which is given by
\begin{equation}
    \label{equation:get_weight}
    \mathbf{W} = \mathrm{Softmax}(\mathbf{S}_m),
\end{equation}
 Given the attention coefficients $\mathbf{W}$, we transform the values $\mathbf{V}$ into refined feature patches $\mathbf{P}\in\mathbb{R}^{N_q\times s\times s\times C^\prime}$, which is given by  
\begin{equation}
    \mathbf{P} = \mathbf{W}\mathbf{V},
\label{eq:refine_feature}
\end{equation}
and use rolling operation \cite{chellapilla:inria-00112631} to form the refined feature maps $\mathbf{F}_{r}$. The temperature parameter $t$ in Eq. (\ref{equation:modify_score}) is adopted to control the output distribution, which corresponds to the attention coefficients used for sum of different neural patches. The lower the temperature $t$, the harder distribution will be, and the higher the temperature $t$, the softer the distribution gets. A harder distribution indicates that the model is more confident to use the most similar neural feature patch, and the model tends to use a greater number of neural feature patches for restoration with a softer distribution. 

\textbf{MHTMA} 
To utilize different degrees of softness and make the model attend on features at multiple distant spatial locations, we design a multi-head masked self-attention (MHTMA) module. Single temperature masked self-attention is believed to attend on one position in each row, since the output of Softmax typically would have one dimension significantly larger than other dimensions in each row. MHTMA module employs multiple temperature scaled masked self-attention in parallel, which was proposed to jointly attend on multiple positions and better capture long-range contextual dependencies. The final refined feature cube $\mathbf{F}_{out}$ of $K$ heads is given by
\begin{equation}
\mathbf{F}_{out}=\mathrm{Concat}_{i\in [K]}[\mathbf{F}_{r}^{i}],
\end{equation}
where $\mathbf{F}_{r}^{i}$ denotes the refined feature cube obtained from the $i$-th head.  As illustrated in Fig. \ref{fig:attention}, the $K$ groups of refined feature cubes are then rolled and concatenated to generate $\mathbf{F}_{out}$. In this work, we design a learning network to adaptively control the temperatures. The details of the temperature learning are presented in the next subsection.

\subsection{Learning-based Self-Adaptive Control of Temperature}
\label{tem}
\begin{figure}[!thb]
    \centering
    \includegraphics[width=0.75\linewidth]{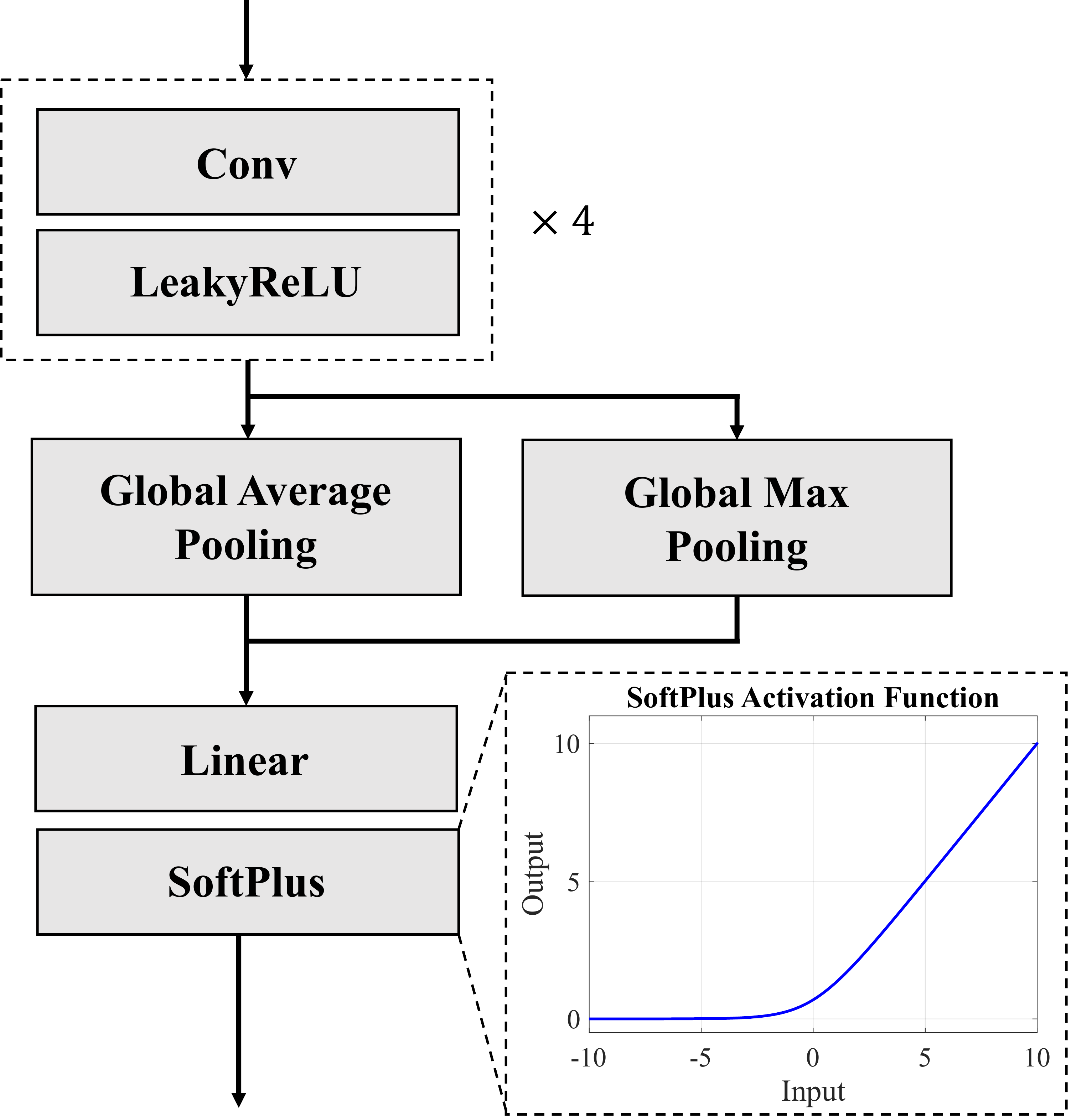}
    \caption{Illustration of our embedding network for temperature learning.}
    \label{fig:embedding}
\end{figure}
We design an embedding network $\mathcal{F}_{emb}$ to learn the temperatures from the input feature maps $\mathbf{F}_{in}$. Inspired by recent CNN-based temperature learning models \cite{lin-etal-2018-learning, zhou2021atma}, our embedding network follows a CNN architecture similar to the one used in \cite{zhou2021atma}. As illustrated in Fig. \ref{fig:embedding}, the  network $\mathcal{F}_{emb}$ first uses 4 convolution layers to encode $\mathbf{F}_{in}$. Then, a global average pooling (GAP) layer and a global max pooling (GMP) layer are employed in parallel to extract global information. After that, the outputs of GAP and GMP are concatenated and transformed by a fully connected layer. In \cite{zhou2021atma}, LeakyReLU was used as the last activation function, which may produce negative temperatures and cause unstable training. To address this issue, our network $\mathcal{F}_{emb}$ uses Softplus as the last activation function, which is given by
\begin{equation}
    \mathrm{Softplus}(x) = \ln(1+e^{-x}),
\end{equation}
and the overall learning of a temperature can be formulated as
\begin{equation}
    t = \mathrm{Softplus}(\mathcal{F}_{emb}(\mathbf{F}_{in})).
\end{equation}

Learned temperature $t$ is then used to scale the attention score $\mathbf{S}_m$ in Eq. (\ref{equation:modify_score}). The training of our attention mechanism using a single learned temperature is given in Algorithm \ref{algorithm:Ours}. For $K$ heads, we use $K$ different channel-wise fully connected networks $\mathcal{F}_{dr}$ to calculate the similarity of patches in projection on different hyperplanes. 

\begin{algorithm}[t!]
    \renewcommand{\algorithmicrequire}{\textbf{Input:}}
    \renewcommand{\algorithmicensure}{\textbf{Output:}}
    \renewcommand{\algorithmiccomment}{\hfill//}
    \caption{Training of the proposed single temperature masked self-attention}
    \label{algorithm:Ours}
    \begin{algorithmic}[1]
        \REQUIRE Input feature map $\mathbf{F}_{in}$, Mask $\mathbf{M}$, Temperature $t$, Batch size $N$.
        \ENSURE Output feature map $\mathbf{F}_{r}$.
        \STATE $\mathbf{F}_{e} \leftarrow \mathcal{F}_{dr}(\mathbf{F}_{in})$.
        \STATE $\mathbf{Q}, \mathbf{K}, \mathbf{V} \leftarrow \mathrm{Unrolling}(\mathbf{F}_{e})$.
        \STATE $\mathbf{S} \leftarrow \mathcal{F}_n(\mathbf{Q}){\mathcal{F}_n(\mathbf{K}})^T$
        \COMMENT{\ Self-attention based.}
        \STATE Compute $\mathbf{M}^\prime$ based on $\mathbf{M}$.
        \STATE $\mathbf{S}_m \leftarrow \mathbf{M}^\prime \odot (\mathbf{S}/t + \lambda_m) - \lambda_m$.
        \STATE $\mathbf{W} \leftarrow \mathrm{Softmax}(\mathbf{S}_m)$.
        \STATE $\mathbf{P} \leftarrow \mathbf{V}\mathbf{W}$.
        \STATE $\mathbf{F}_{r}\leftarrow \mathrm{Rolling}(\mathbf{P})$.
    \end{algorithmic}
\end{algorithm}

\section{Experiments}
\subsection{Datasets}
We evaluate the proposed image inpainting method on three benchmark datasets including Paris StreetView \cite{doersch2012what}, CelebA-HQ \cite{DBLP:conf/iclr/KarrasALL18} and Places2 \cite{zhou2017places}.

\textbf{Paris StreetView} \cite{doersch2012what}: It contains 15,000 images with resolution of 936$\times$537. The images are mainly composed of highly structured buildings with regular patterns, supplemented by urban natural landscapes. We follow the default split that produces 14,900 training images and 100 validation images.

\textbf{CelebA-HQ} \cite{DBLP:conf/iclr/KarrasALL18}: It contains 30,000 aligned face images with a maximum resolution of 1024$\times$1024. We randomly select 1,000 male and 1,000 female faces to form a gender-balanced test set, and use the remaining 28,000 for training.

\textbf{Places2} \cite{zhou2017places}: We use the high-resolution version of Places365-Standard dataset as our Places2 dataset. It contains 1,803,460 training samples of more than 300 scenes. We randomly select 2000 images from the validation set as test images in our experiments. 

To generate incomplete images as inputs, we use a mask generation method presented in \cite{Yu_2019_ICCV}. Specifically, a union of a random free-form mask and a 96$\times$96 sized square mask is generated on-the-fly during training and inference.

\subsection{Implementation details}
The proposed model is trained with PyTorch v1.9.1, CUDA v11.1, cuDNN v8.0.5 on a PC with an Intel Xeon CPU and two 24G NVIDIA GeForce RTX 3090 GPUs. The resolution of the input images is 256$\times$256. Before feeding the images into the model, we first resize the images and augment them using horizontal flip with a probability of 0.5. The missing regions are initialized with a constant value, i.e., ImageNet mean RGB value $[0.485, 0.456, 0.406]$. The proposed model is trained with a batch size of 16 and optimized using Adam optimizer \cite{DBLP:journals/corr/KingmaB14} with $\beta_1=0.5$ and $\beta_2=0.9$. The learning rates for the generator and discriminators are set to 1e-4 and 1e-4, respectively. The training process is terminated when the validation loss is not decreased within 30 epochs, and the model with the smallest validation loss is saved for test. It takes around 6 days for training Paris StreetView model, 7 days for CelebA-HQ and 10 days for Places2. The inference is performed on a single GPU, and our model takes 0.1 seconds per 256$\times$256 image.




\begin{figure*}[tbh]
    \centering
    \null\hfill
    \subfloat[Ground Truth]{\includegraphics[width=0.18\linewidth]{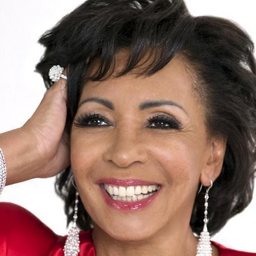}}\hfill
    \subfloat[Input]{\includegraphics[width=0.18\linewidth]{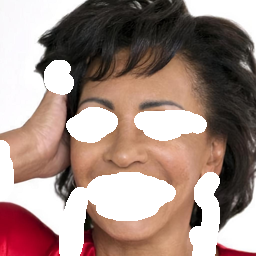}}\hfill
    \subfloat[W/ CA]{\includegraphics[width=0.18\linewidth]{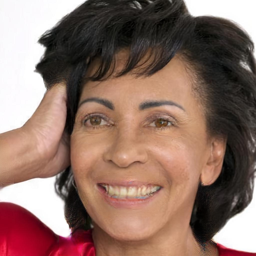}}\hfill
    \subfloat[W/ ATMA]{\includegraphics[width=0.18\linewidth]{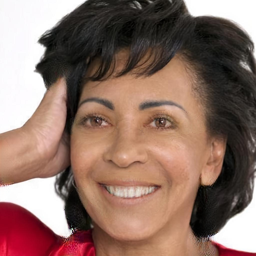}}\hfill
    \subfloat[W/ MHTMA]{\includegraphics[width=0.18\linewidth]{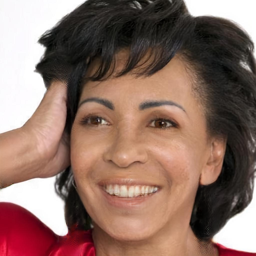}}
    \hfill\null
    \caption{Visual comparison between the proposed model with MHTMA and the two baseline models, which replace the MHTMA with contextual attention and ATMA. The MHTMA-based model generates more natural results with fewer artifacts than the CA-based and ATMA-based baseline models.}
    \label{fig:attentions}
\end{figure*}

\subsection{Experiment setup}
We conduct three experiments to analyze and evaluate the proposed image inpainting model. First, we do ablation studies to analyze the impact of different temperature scaling methods, different generation branches in the refinement network, and different adversarial loss functions on inpainting quality. Then, we do quantitative and qualitative comparisons between the proposed model and five existing state-of-the-art image inpainting methods:

-\ PatchMatch \cite{10.1145/1531326.1531330}: It is known as one of the state-of-the-art non-learning-based approaches. 
    
-\ ParConv \cite{10.1007/978-3-030-01252-6_6}: It is a partial convolution-based image inpainting model for irregular holes.  
    
-\ DeepFillv2 \cite{Yu_2019_ICCV}: It is an improved version of the image inpainting method DeepFill \cite{8578675}, by combining gated convolution generator along with spectral-normalized discriminator to handle free-form image inpainting. 
    
-\ WaveFill \cite{yu2021wavefill}: It is a wavelet-based image inpainting method, which synthesizes missing regions at different frequency bands explicitly and separately. 
    
-\ ATMA \cite{zhou2021atma}: It is a free-form image inpainting method, which includes an adaptive multi-temperature mask-guided attention mechanism in the refinement network of DeepFill \cite{8578675}.

\begin{table}[ht!]
    \caption{Quantitative Comparison of Image Inpainting Using Different Attention Mechanisms in Networks on CelebA-HQ}
    \centering
    \begin{tabular}{l|llll}
    \hline\hline
    Methods & MAE $\downarrow$ & PSNR $\uparrow$ & SSIM $\uparrow$ & FID $\downarrow$   \\ \hline
    ours & \textbf{1.282\%} & \textbf{28.82} & \textbf{0.9167} & \textbf{5.147}   \\
    w/ ATMA & 1.372\% & 28.44 & 0.9108 & 5.799   \\
    w/ CA & 1.309\% & 28.71 & 0.9132 & 5.425  \\
    \hline\hline
    \end{tabular}
    \label{table:ablation_baseline}
\end{table}

We use the Content-Aware Fill of Adobe Photoshop to generate the inpainting results of PatchMatch, and retrain all deep learning models including ParConv\footnote[1]{ParConv: https://github.com/naoto0804/pytorch-inpainting-with-partial-conv}, DeepFillv2\footnote[2]{DeepFillv2: https://github.com/JiahuiYu/generative\_inpainting}, WaveFill\footnote[3]{WaveFill: https://github.com/yingchen001/WaveFill} and ATMA using exactly the same dataset setting and mask generation method as our model. After that, we evaluate the effectiveness of the proposed model in user-guided inpainting.

\subsection{Ablation studies}
\textbf{Effect of MHTMA} To evaluate the effectiveness of the proposed MHTMA module, we compare our model with two baseline free-form inpainting models: ATMA-based model and CA-based model. The two baseline models are modified versions of the proposed model, where ATMA-based model replaces MHTMA with ATMA proposed in \cite{zhou2021atma} and the CA-based model replaces MHTMA with the Contextual Attention (CA) proposed in \cite{Yu_2019_ICCV}. The contextual attention mechanism is a convolution-based self-attention algorithm, which uses a constant temperature to adjust the attention scores and generate finer details. The training of the contextual attention module is given in Algorithm \ref{algorithm:CA}.

\begin{algorithm}[thb]
    \renewcommand{\algorithmicrequire}{\textbf{Input:}}
    \renewcommand{\algorithmicensure}{\textbf{Output:}}
    \renewcommand{\algorithmiccomment}{\ //}
    \caption{Training of contextual attention module \cite{8578675}}
    \label{algorithm:CA}
    \begin{algorithmic}[1]
        \REQUIRE Input feature map $\mathbf{F}_{in}$, Mask $\mathbf{M}$, Temperature $\frac{1}{10}$, Batch size $N$.
        \ENSURE Output feature map $\mathbf{F}_{out}$.
        \FOR{$n=1\dots N$}
            \STATE $\mathbf{p}_n \leftarrow$ $\mathrm{Unrolling}(\mathbf{F}_{in, n})$.
            \STATE $\mathbf{p}_n^{\prime} \leftarrow\mathcal{F}_n(\mathbf{p}_n)$.
            \COMMENT{\ Normalization.}
            \STATE $\mathbf{S}_n \leftarrow\mathrm{Conv2d}(\mathbf{F}_{in, n}, \mathbf{p}_n^{\prime})$.
            \COMMENT{\ Convolution-based.}
            \STATE Compute $\mathbf{M}_n^\prime$ according to $\mathbf{M}_n$.
            \STATE $\mathbf{S}_{m,n} \leftarrow \mathbf{M}_n^\prime\odot\mathbf{S}_n$.
            \STATE $\mathbf{W}_n \leftarrow \mathrm{Softmax}(\mathbf{S}_{m, n}/\frac{1}{10})\odot\mathbf{M}_n^\prime$.
            \STATE $\mathbf{F}_{out, n} \leftarrow \mathrm{TransposedConv2d}(\mathbf{W}_n, \mathbf{p}_n)$.
        \ENDFOR
        \STATE $\mathbf{F}_{out} \leftarrow \left(\mathbf{F}_{out, 1}, \mathbf{F}_{out, 2}, \dots, \dots, \mathbf{F}_{out, N}\right)$.
    \end{algorithmic}
\end{algorithm}

The visual comparison results in Fig. \ref{fig:attentions} and the qualitative comparison results in Table \ref{table:ablation_baseline} show that the proposed model with MHTMA generates inpainting results with better visual and numerical performance than the two baseline models. This is because that MHTMA can learn temperatures in a stable way and uses suitable neural patches for high quality image generation. In addition, since attention scores in ATMA are adjusted using learned temperatures, the ATMA-based model generates more realistic results with fewer artifacts than the CA-based model.

\begin{figure}[bth!]
    \centering
    \null\hfill
    \subfloat[Our model]{\includegraphics[width=0.48\linewidth]{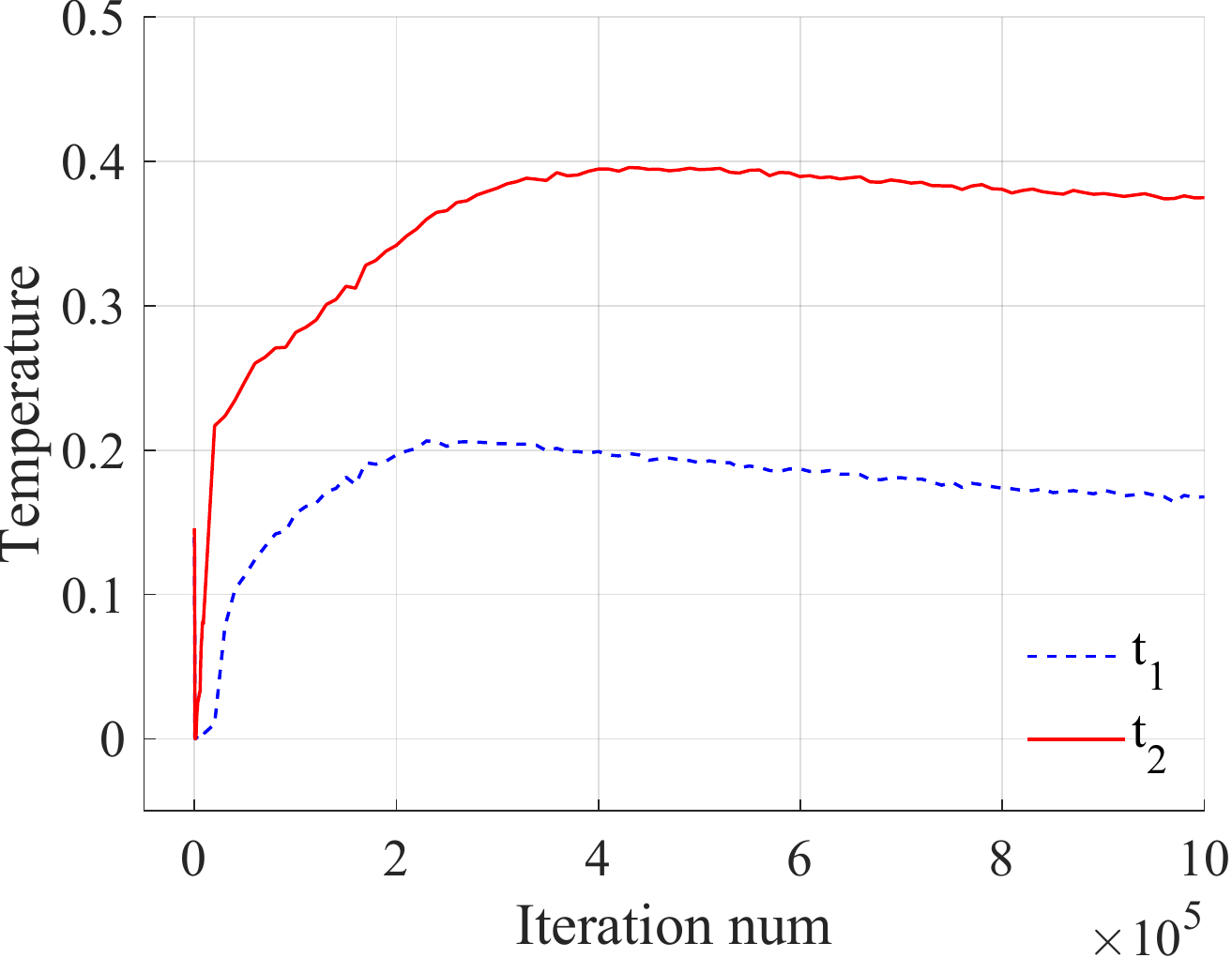}
    \label{subfig:train_t_ours}}
    \hfill
    \subfloat[ATMA-based model]{\includegraphics[width=0.48\linewidth]{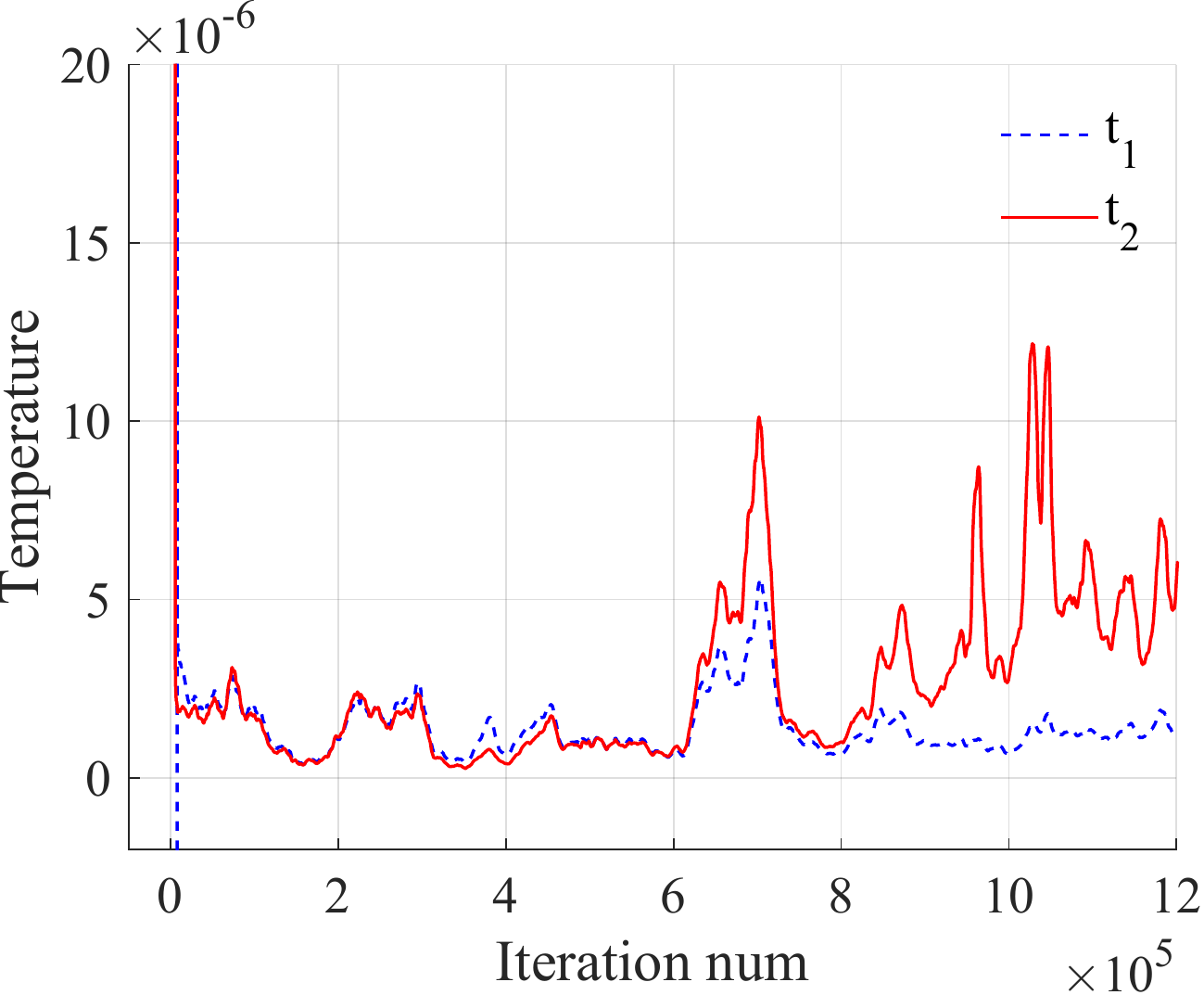}
    \label{subfig:train_t_ATMA}}
    \hfill\null
    \caption{Temperature training curves of the proposed model and ATMA-based model. Replacing activation function LeakyReLU with Softplus makes the temperature learning network converge to a high temperature and a low temperature.}
    \label{fig:train_t}
\end{figure}

\textbf{Temperature training curves} The temperature training curves of the proposed model and ATMA-based model\ \cite{zhou2021atma} are illustrated in Fig. \ref{fig:train_t}\subref{subfig:train_t_ours} and Fig. \ref{fig:train_t}\subref{subfig:train_t_ATMA}, respectively. We observe that the proposed model learns two different temperatures $t_{1}$ and $t_2$. The temperature learning process of our model can be divided into three stages. At the beginning of the training, temperatures cool rapidly to copy-and-paste the nearest neighbor. This can be explained that the nearest neighbor patch provides more meaningful information for image restoration than learned features at the beginning of the training. Since the learned features gradually contain meaningful information after the model is trained for a period of time, temperatures warm up to make use of more similar patches for image restoration. The model converges to a high temperature and a low temperature with enough training iterations. In addition, although ATMA-based model learns two different temperatures with enough training iterations, the learning process is unstable and fails to converge.  This is because that ATMA uses LeakyReLU as the last activation function for temperature learning, allowing negative temperature and causing unstable training. Unlike ATMA, the proposed MHTMA uses Softplus to restrict the learned temperatures to be positive and gets stable temperature learning.    

\begin{figure}[htb!]
    \centering
    \null\hfill
    \subfloat[On single RTX 3090 GPU]{\includegraphics[width=0.48\linewidth]{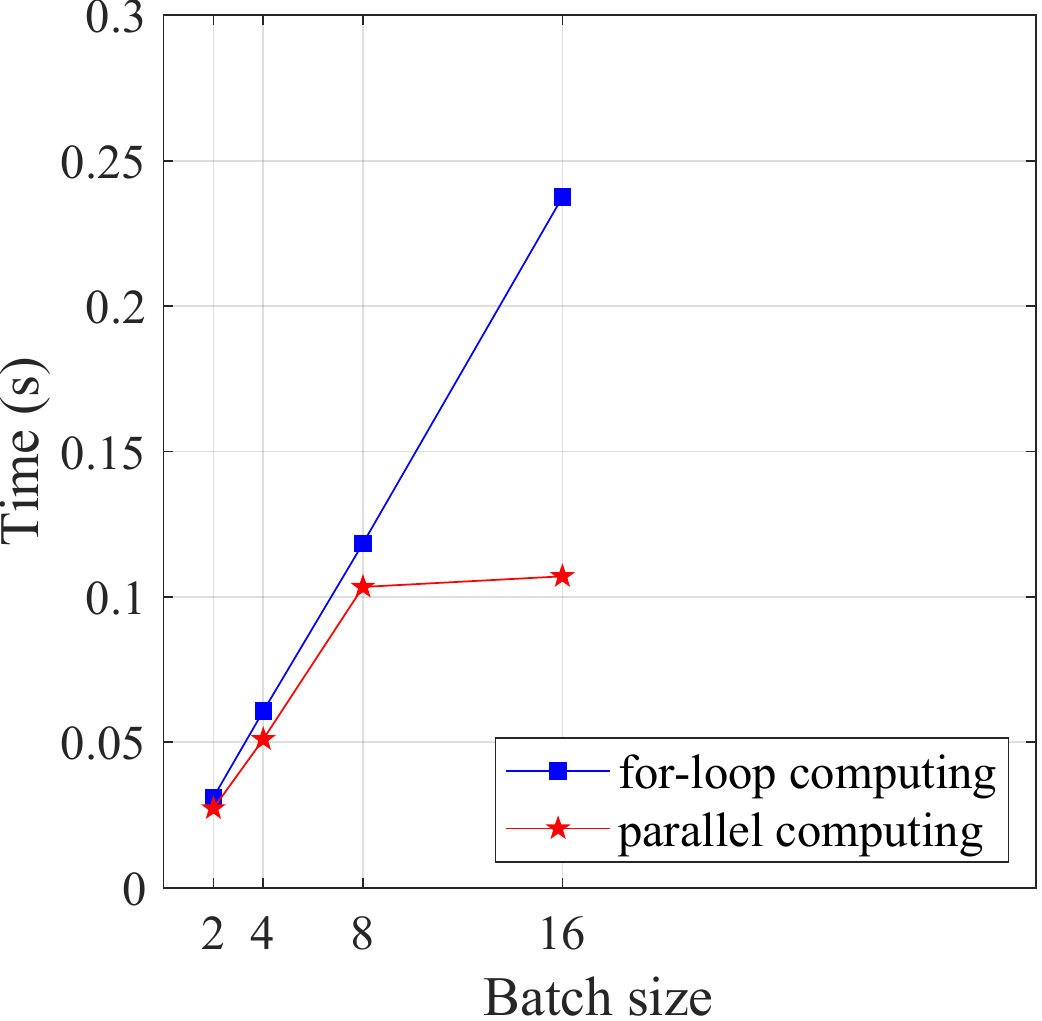}
    \label{subfig:oneGPU}}
    \hfill
    \subfloat[On two RTX 3090 GPUs]{\includegraphics[width=0.48\linewidth]{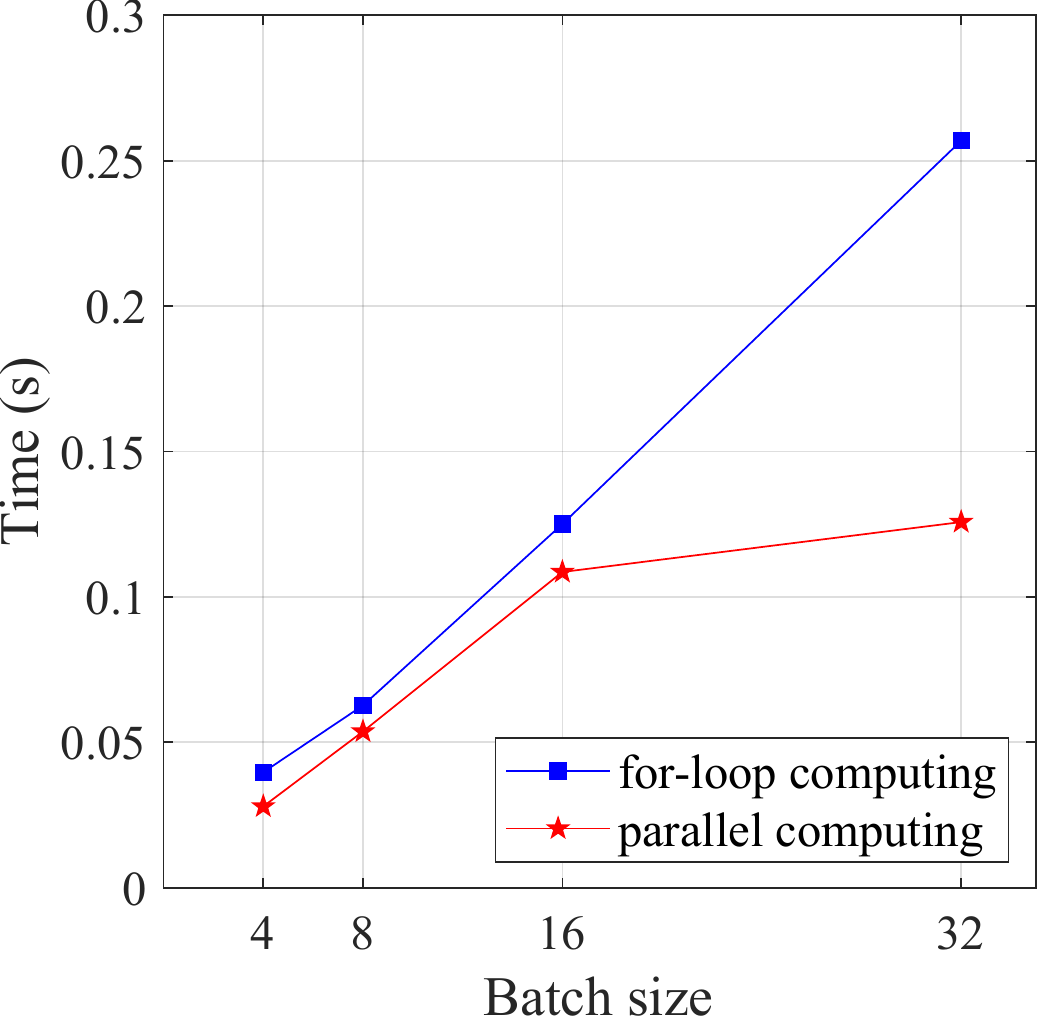}
    \label{subfig:twoGPU}}
    \hfill\null
    \caption{Comparison of computational efficiency between for-loop and parallel implementation of our attention. We observe that the proposed model with parallel computing consumes less time than a baseline model using existing for-loop computing under different batch sizes.}
    \label{fig:speed}
\end{figure}

\textbf{Computational efficiency of the proposed model} We compare the computational efficiency of our model with those of our previous model ATMA \cite{zhou2021atma}. We repeat the feedforward pass of each model 500 times and calculate their average consuming time. Fig. \ref{fig:speed}\subref{subfig:oneGPU} and Fig. \ref{fig:speed}\subref{subfig:twoGPU} show the comparison results on single and two GPUs, respectively. The results in Fig. \ref{fig:speed} show that the time for ATMA to make a feedforward pass is increased linearly with increasing the number of batch size. Compared with ATMA, the proposed model consumes less time under different batch sizes, especially with 16 batch size per GPU.

\begin{figure*}[!h]
    \centering
    \setlength\tabcolsep{1pt}
    \captionsetup{justification=centering}
    \begin{tabular}{ccccccc}
        \includegraphics[width=0.14\linewidth]{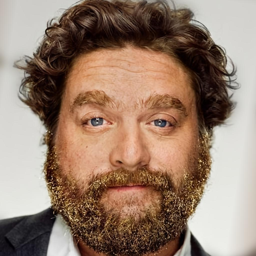} & \includegraphics[width=0.14\linewidth]{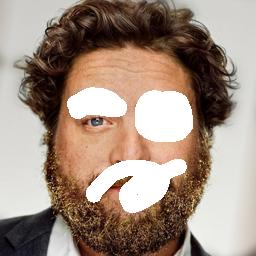} & \includegraphics[width=0.14\linewidth]{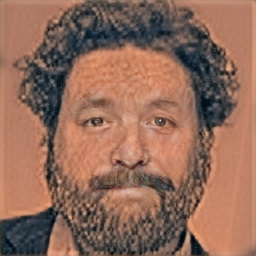} & \includegraphics[width=0.14\linewidth]{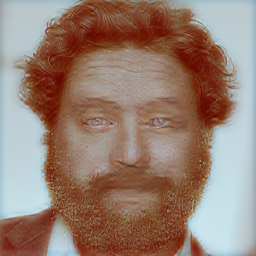} & \includegraphics[width=0.14\linewidth]{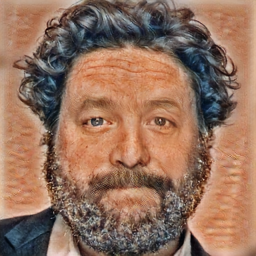} & \includegraphics[width=0.14\linewidth]{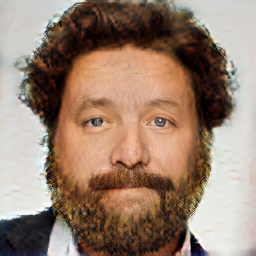} & \includegraphics[width=0.14\linewidth]{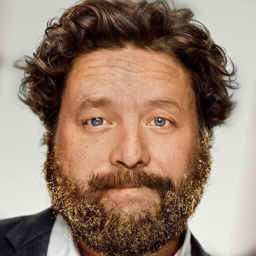} \\
        \subfloat[Ground Truth]{\includegraphics[width=0.14\linewidth]{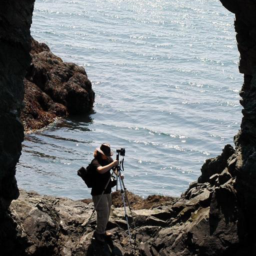}\label{subfig:branches_gt}} & \subfloat[Input]{\includegraphics[width=0.14\linewidth]{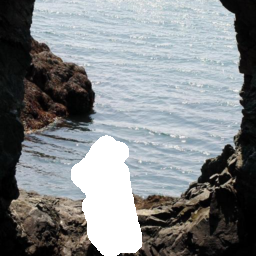}\label{fig:branches_in}} & \subfloat[W/ convolution]{\includegraphics[width=0.14\linewidth]{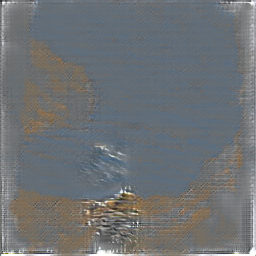}\label{subfig:branches_conv}} & \subfloat[W/ attention]{\includegraphics[width=0.14\linewidth]{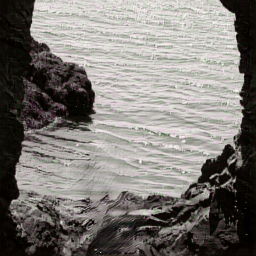}\label{subfig:branches_attention}} & \subfloat[W/o low temperature attention]{\includegraphics[width=0.14\linewidth]{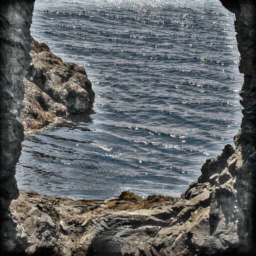}\label{subfig:branches_high}} & \subfloat[W/o high temperature attention]{\includegraphics[width=0.14\linewidth]{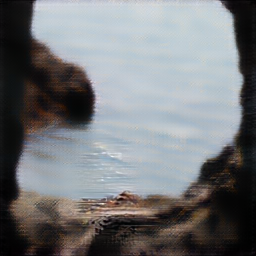}\label{subfig:branches_low}} & \subfloat[Full Model]{\includegraphics[width=0.14\linewidth]{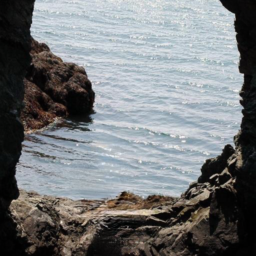}\label{subfig:branches_full}}
    \end{tabular}
    \caption{Image inpainting results of different branches used in the second stage. The inpainting results in (c) and (d) show that either convolution or attention branch can generate reasonable semantic information for missing regions, and  generate different texture details.  Results in (e) and (f) show that our model with low temperature head in attention branch generates more visible acceptable results than the one with high temperature head. This indicates that the low temperature branch helps to generate the certain and basic part of image, while high temperature branch helps the model to generate the uncertain and fine part of image. Inpainting results of our full model in (g) show that two temperatures help the model capture more long-range contextual information for high quality image inpainting.}
    \label{fig:branches}
\end{figure*}
\textbf{Effects of different branches in the second stage of the proposed model} The second stage of our model consists of two branches: a convolution branch and an attention branch, for generating finer details based on the first stage. Existing two-stage image inpainting models \cite{Yu_2019_ICCV, zhou2021atma} have demonstrated the effectiveness of the dual-branch-based second stage on finer detail generation. In this subsection, we analyze how each branch affects the feature extraction and image generation of the proposed model, e.g., turning off the attention branch during inference to visualize the image inpainting results of our model. Fig. \ref{fig:branches} shows the inpainting results of the proposed model with different branches in the second stage. Fig. \ref{fig:branches}\subref{subfig:branches_conv} and Fig. \ref{fig:branches}\subref{subfig:branches_attention} show that both convolution branch and attention branch can restore the missing regions with reasonable semantic information, but with different texture information. In addition, Fig. \ref{fig:branches}\subref{subfig:branches_high} and Fig. \ref{fig:branches}\subref{subfig:branches_low} show  the inpainting results of our model using high temperature and low temperature in attention branch, respectively. We observe that the inpainting results of the model with high temperature in attention branch contain more texture details, but suffer from uncertainty and inconsistency in color and texture generation. The model with low temperature in attention branch generates more realistic inpainting results with certain and consistent colors and textures. 
Our full model with both convolution and attention branches produces the final visually realistic results, see Fig. \ref{fig:branches}\subref{subfig:branches_full}.

\textbf{WGAN-GP vs HAL-SN} Wasserstein Generative Adversarial Network (WGAN-GP) \cite{NIPS2017_892c3b1c} and Hinge Adversarial Loss with discriminator Spectral Normalization (HAL-SN) \cite{DBLP:journals/corr/abs-1802-05957, lim2017geometric} are two general methods for stabilizing training in GAN-based image generation. We do ablation studies on CelebA-HQ to illustrate the effectiveness of WGAN-GP and HAL-SN. Quantitative comparison results in Table \ref{table:adv_loss_compare} show that our model with HAL-SN generates slightly better inpainting results than the one with WGAN-GP. In addition, the training of the HAL-SN-based model is faster than those of the WGAN-GP-based model, since the HAL-SN-based model processes 71.8 batches per minute and the WGAN-GP-based model processes 59.0 batches per minute. Thus, we use HAL-SN loss instead of WGAN-GP loss in our model.

\begin{table}[!h]
    \centering
    \scriptsize
    \caption{Quantitative Comparison of WGAN-GP and HAL-SN on CelebA-HQ.}
    \resizebox{0.95\linewidth}{!}{
        \begin{tabular}{l|lcll}
            \hline\hline
            & MAE $\downarrow$ & PSNR(dB) $\uparrow$ & SSIM $\uparrow$ & FID $\downarrow$ \\
            \hline
            WGAN-GP & 1.603\%  & 26.39 & 0.8981 & 5.559\\
            HAL-SN & \textbf{1.571\%} & \textbf{26.67} & \textbf{0.9008} & \textbf{5.392}\\
            \hline\hline
        \end{tabular}
    }
    \label{table:adv_loss_compare}
\end{table}

\subsection{Comparison with existing work}
\begin{figure*}[!h]
    \centering
    \setlength\tabcolsep{1pt}
    \begin{tabular}{cccccccc}
        \includegraphics[width=0.12\linewidth]{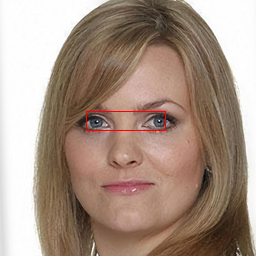} & \includegraphics[width=0.12\linewidth]{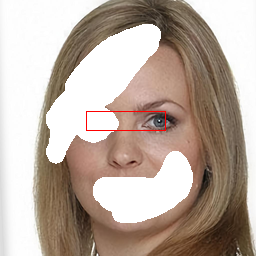} & \includegraphics[width=0.12\linewidth]{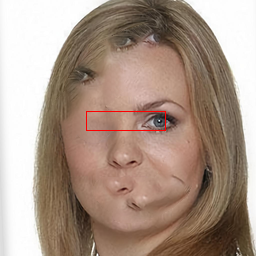} & \includegraphics[width=0.12\linewidth]{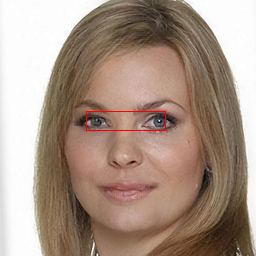} & \includegraphics[width=0.12\linewidth]{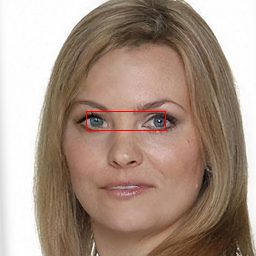} & \includegraphics[width=0.12\linewidth]{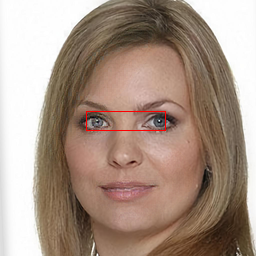} & \includegraphics[width=0.12\linewidth]{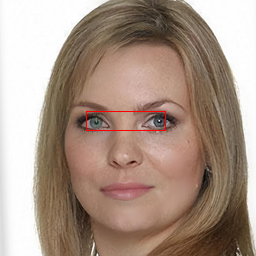} & \includegraphics[width=0.12\linewidth]{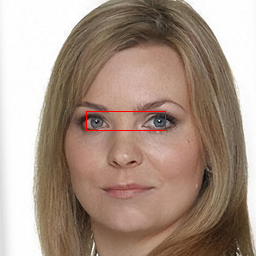} \\[-3pt]
        \includegraphics[width=0.12\linewidth]{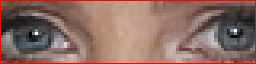} & \includegraphics[width=0.12\linewidth]{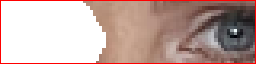} & \includegraphics[width=0.12\linewidth]{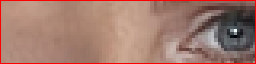} & \includegraphics[width=0.12\linewidth]{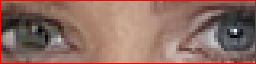} & \includegraphics[width=0.12\linewidth]{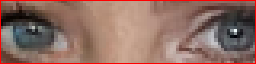} & \includegraphics[width=0.12\linewidth]{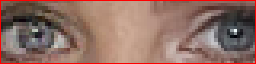} & \includegraphics[width=0.12\linewidth]{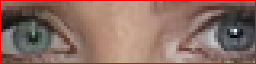} & \includegraphics[width=0.12\linewidth]{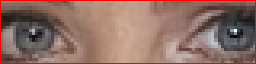} \\[0pt]

        \includegraphics[width=0.12\linewidth]{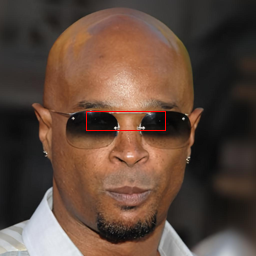} & \includegraphics[width=0.12\linewidth]{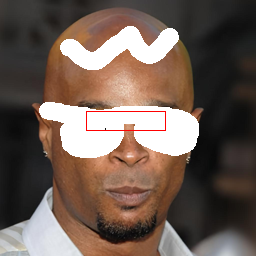} & \includegraphics[width=0.12\linewidth]{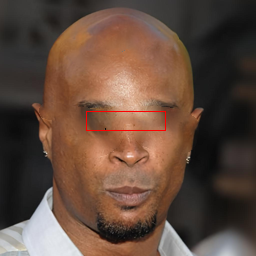} & \includegraphics[width=0.12\linewidth]{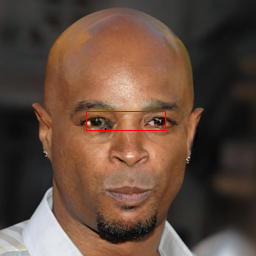} & \includegraphics[width=0.12\linewidth]{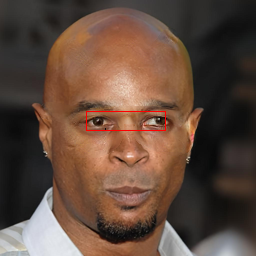} & \includegraphics[width=0.12\linewidth]{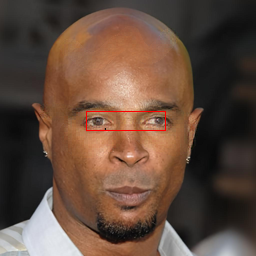} & \includegraphics[width=0.12\linewidth]{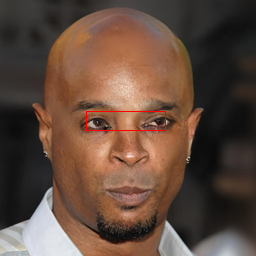} & \includegraphics[width=0.12\linewidth]{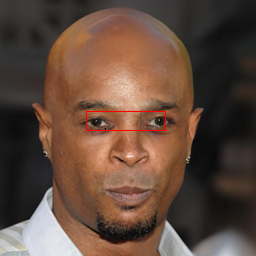} \\[-3pt]
        \includegraphics[width=0.12\linewidth]{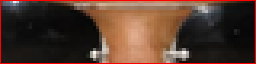} & \includegraphics[width=0.12\linewidth]{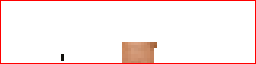} & \includegraphics[width=0.12\linewidth]{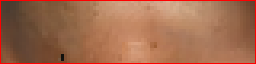} & \includegraphics[width=0.12\linewidth]{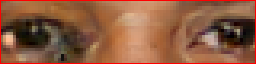} & \includegraphics[width=0.12\linewidth]{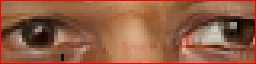} & \includegraphics[width=0.12\linewidth]{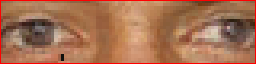} & \includegraphics[width=0.12\linewidth]{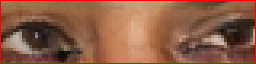} & \includegraphics[width=0.12\linewidth]{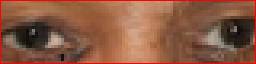} \\[0pt]

        \includegraphics[width=0.12\linewidth]{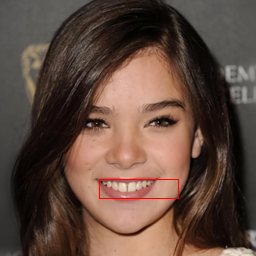} & \includegraphics[width=0.12\linewidth]{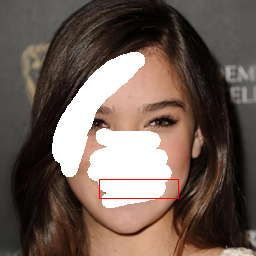} & \includegraphics[width=0.12\linewidth]{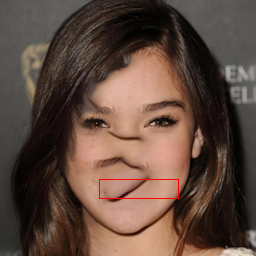} & \includegraphics[width=0.12\linewidth]{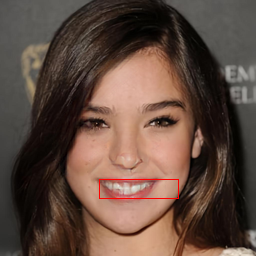} & \includegraphics[width=0.12\linewidth]{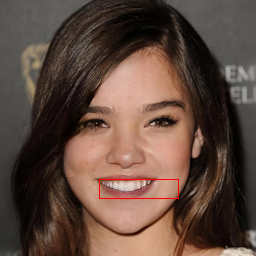} & \includegraphics[width=0.12\linewidth]{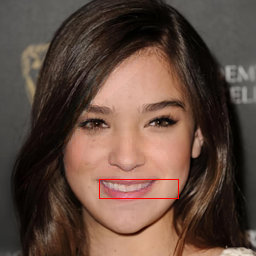} & \includegraphics[width=0.12\linewidth]{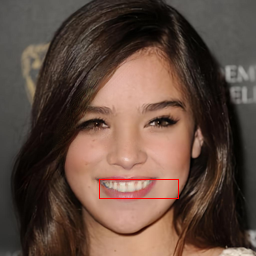} & \includegraphics[width=0.12\linewidth]{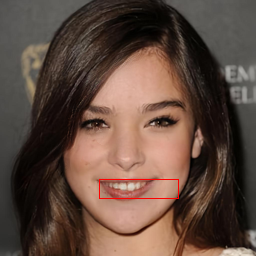} \\[-3pt]
        \subfloat[Ground Truth]{\includegraphics[width=0.12\linewidth]{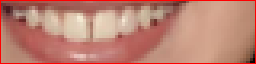}} & \subfloat[Input]{\includegraphics[width=0.12\linewidth]{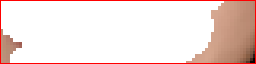}} & \subfloat[PatchMatch\ \cite{10.1145/1531326.1531330}]{\includegraphics[width=0.12\linewidth]{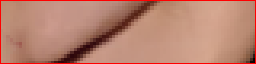}} & \subfloat[ParConv\ \cite{10.1007/978-3-030-01252-6_6}]{\includegraphics[width=0.12\linewidth]{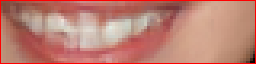}} & \subfloat[DeepFillv2\ \cite{Yu_2019_ICCV}]{\includegraphics[width=0.12\linewidth]{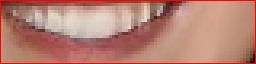}} & \subfloat[WaveFill\ \cite{yu2021wavefill}]{\includegraphics[width=0.12\linewidth]{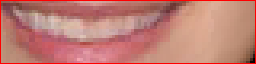}} & \subfloat[ATMA\ \cite{zhou2021atma}]{\includegraphics[width=0.12\linewidth]{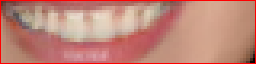}} & \subfloat[Ours]{\includegraphics[width=0.12\linewidth]{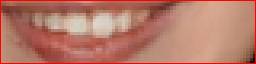}} \\[0pt]
    \end{tabular}
    \caption{Qualitatively comparison results on CelebA-HQ dataset. Our model generates more visually pleasant results with finer details, such as details in eyes and teeth. Zoom in for better visualization.}
    \label{fig:qualitative_celebahq}
\end{figure*}

\textbf{Qualitative comparison} Fig. \ref{fig:qualitative_celebahq} shows qualitative comparison results on CelebA-HQ. We observe that PatchMatch fails to recover large missing regions with complex content by copying similar patches from known regions. ParConv generates reasonable structures with blurry textures. DeepFillv2 makes improvements by gated convolution and contextual attention, but still exhibits observable unpleasant boundaries and blurry textures. WaveFill generates more semantically reasonable results with less artifacts. However, some restored regions still contain unnatural texture details, such as eyes and mouths. ATMA produces results with more texture details by capturing long-range contextual information with learnable temperatures, but the generated details in teeth and mouths are still not as delicate as the known ones. As a comparison, the inpainting results by the proposed method are more natural with finer textures and less color inconsistency.  

\begin{figure*}[!h]
    \centering
    \setlength\tabcolsep{1pt}
    \begin{tabular}{cccccccc}
        \includegraphics[width=0.12\linewidth]{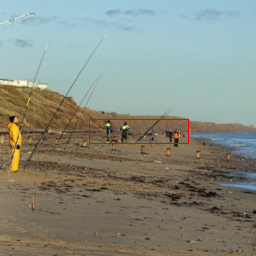} & \includegraphics[width=0.12\linewidth]{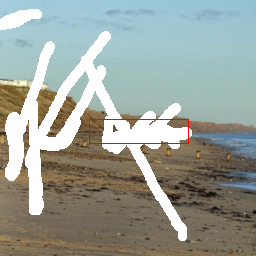} & \includegraphics[width=0.12\linewidth]{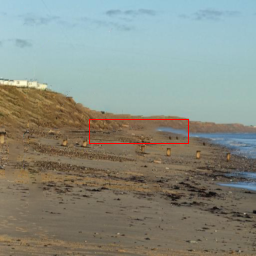} & \includegraphics[width=0.12\linewidth]{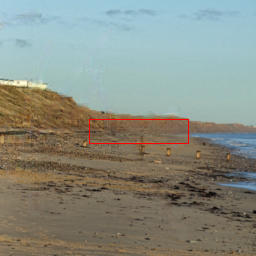} & \includegraphics[width=0.12\linewidth]{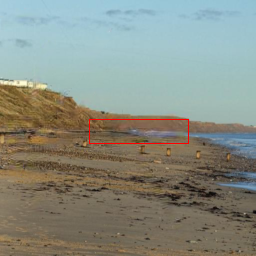} & \includegraphics[width=0.12\linewidth]{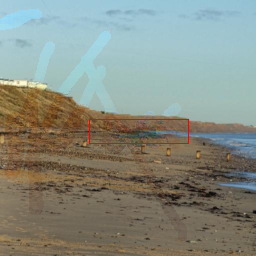} & \includegraphics[width=0.12\linewidth]{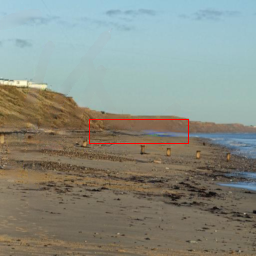} & \includegraphics[width=0.12\linewidth]{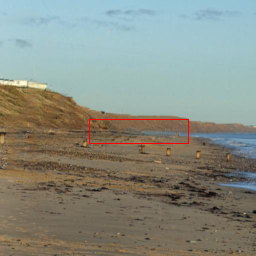} \\[-3pt]
        \includegraphics[width=0.12\linewidth]{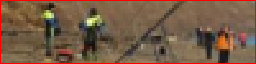} & \includegraphics[width=0.12\linewidth]{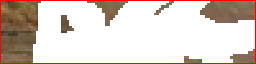} & \includegraphics[width=0.12\linewidth]{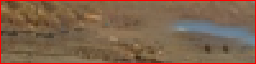} & \includegraphics[width=0.12\linewidth]{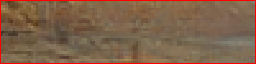} & \includegraphics[width=0.12\linewidth]{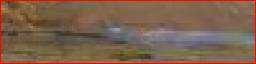} & \includegraphics[width=0.12\linewidth]{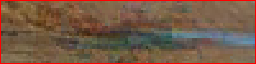} & \includegraphics[width=0.12\linewidth]{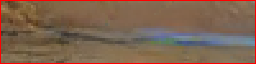} & \includegraphics[width=0.12\linewidth]{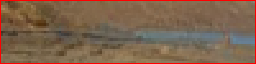} \\[0pt]

        \includegraphics[width=0.12\linewidth]{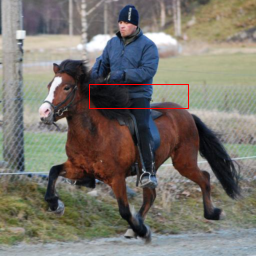} & \includegraphics[width=0.12\linewidth]{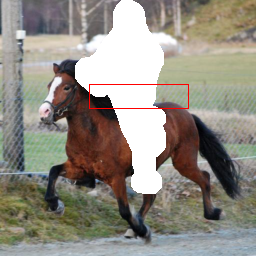} & \includegraphics[width=0.12\linewidth]{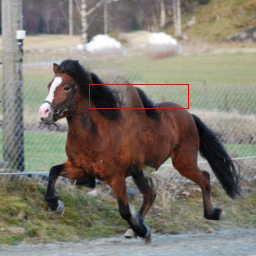} & \includegraphics[width=0.12\linewidth]{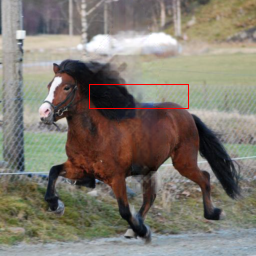} & \includegraphics[width=0.12\linewidth]{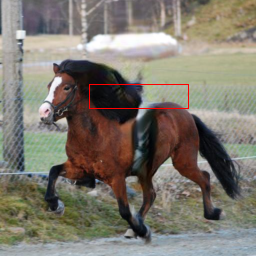} & \includegraphics[width=0.12\linewidth]{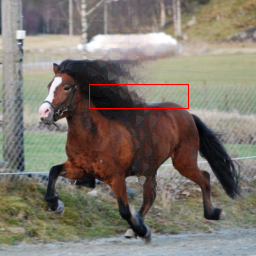} & \includegraphics[width=0.12\linewidth]{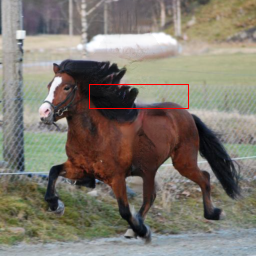} & \includegraphics[width=0.12\linewidth]{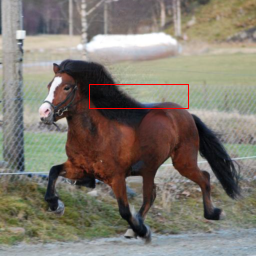} \\[-3pt]
        \includegraphics[width=0.12\linewidth]{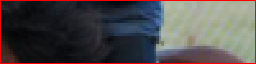} & \includegraphics[width=0.12\linewidth]{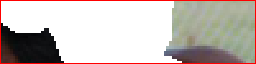} & \includegraphics[width=0.12\linewidth]{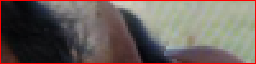} & \includegraphics[width=0.12\linewidth]{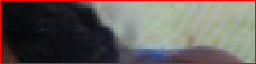} & \includegraphics[width=0.12\linewidth]{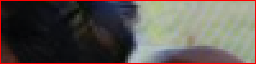} & \includegraphics[width=0.12\linewidth]{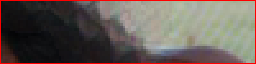} & \includegraphics[width=0.12\linewidth]{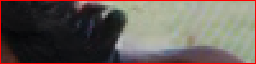} & \includegraphics[width=0.12\linewidth]{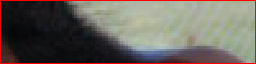} \\[0pt]

        \includegraphics[width=0.12\linewidth]{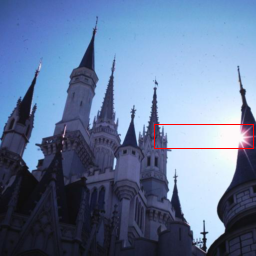} & \includegraphics[width=0.12\linewidth]{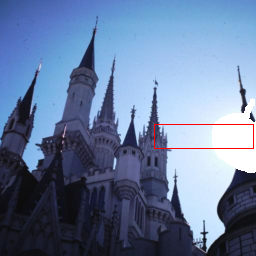} & \includegraphics[width=0.12\linewidth]{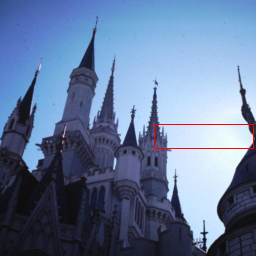} & \includegraphics[width=0.12\linewidth]{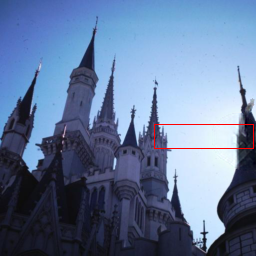} & \includegraphics[width=0.12\linewidth]{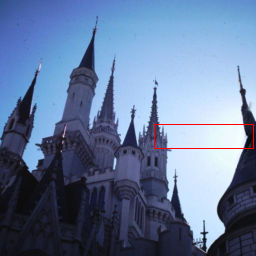} & \includegraphics[width=0.12\linewidth]{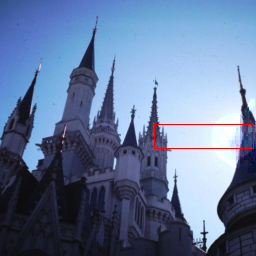} & \includegraphics[width=0.12\linewidth]{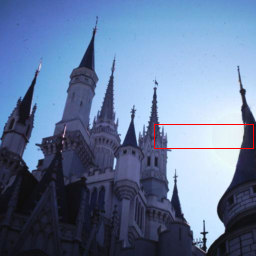} & \includegraphics[width=0.12\linewidth]{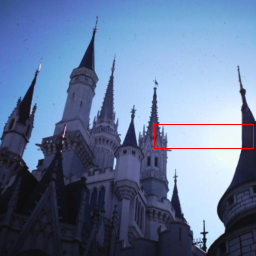} \\[-3pt]
        \subfloat[Ground Truth]{\includegraphics[width=0.12\linewidth]{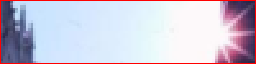}} & \subfloat[Input]{\includegraphics[width=0.12\linewidth]{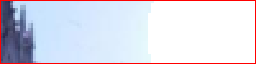}} & \subfloat[PatchMatch\ \cite{10.1145/1531326.1531330}]{\includegraphics[width=0.12\linewidth]{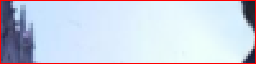}} & \subfloat[ParConv\ \cite{10.1007/978-3-030-01252-6_6}]{\includegraphics[width=0.12\linewidth]{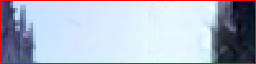}} & \subfloat[DeepFillv2\ \cite{Yu_2019_ICCV}]{\includegraphics[width=0.12\linewidth]{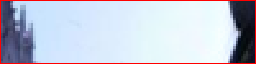}} & \subfloat[WaveFill\ \cite{yu2021wavefill}]{\includegraphics[width=0.12\linewidth]{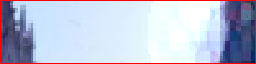}} & \subfloat[ATMA\ \cite{zhou2021atma}]{\includegraphics[width=0.12\linewidth]{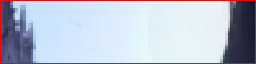}} & \subfloat[Ours]{\includegraphics[width=0.12\linewidth]{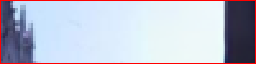}} \\[0pt]
    \end{tabular}
    \caption{Qualitatively comparison results on Places2 dataset. Our model generates more reasonable object and scene structures with better color consistency and fewer artifacts. Zoom in for better visualization.}
    \label{fig:qualitative_places2}
\end{figure*}

Next, we report the visual comparison results on the more challenging Places2 dataset by comparing to the previous PatchMatch \cite{10.1145/1531326.1531330}, DeepFillv2 \cite{Yu_2019_ICCV}, WaveFill \cite{yu2021wavefill} and ATMA \cite{zhou2021atma}. As shown in Fig. \ref{fig:qualitative_places2}, PatchMatch works well on this dataset with repetitive structures. DeepFillv2 and WaveFill generate reasonable completion results but still contain distorted structures and texture artifacts. ATMA uses more similar features in attention-based finer inpainting stage and improves inpainting performance with fewer texture artifacts, but it has noticeable color inconsistency near edge regions when the mask regions contain colorful contents. Since our model uses improved network architecture and attention module, it obtains more visually pleasing results with more semantic structures and better texture details than the other inpainting approaches.

\begin{table}
    \centering
    \caption{Quantitative comparison on validation images of CelebA-HQ with random generated free-form masks.}
    \scriptsize
    \resizebox{0.95\linewidth}{!}{
        \begin{tabular}{l|lcll}
            \hline\hline
            &MAE $\downarrow$ & PSNR(dB) $\uparrow$ & SSIM $\uparrow$ & FID $\downarrow$ \\ 
            \hline
            ParConv\ \cite{10.1007/978-3-030-01252-6_6} & 1.554\% & 27.70 & 0.9000 & 7.606 \\
            DeepFillv2\ \cite{Yu_2019_ICCV} & 1.680\% & 27.24 & 0.8973 & 7.864 \\
            WaveFill\ \cite{yu2021wavefill} & 1.541\% & 27.82 & 0.9059 & 6.968 \\ 
            ATMA\ \cite{zhou2021atma} & 1.297\% & 28.71 & 0.9161 & 5.692 \\
            ours & \textbf{1.282\%} & \textbf{28.82} & \textbf{0.9167} & \textbf{5.147} \\
            \hline\hline
        \end{tabular}
    }
    \label{table:quantitative_celebahq}
\end{table}

\begin{table}[tbh!]
    \caption{Quantitative comparison on validation images of Places2 with masks under different mask ratios.}
    \centering
    \scriptsize
    \begin{tabular}{l|l|l|l|l|l|l}
        \hline\hline
                      & Ratio & ParConv & DeepFillv2 & WaveFill & ATMA & ours \\ \hline
    \multirow{5}{*}{\rotatebox{90}{MAE $\downarrow$}} &
                        (0.0, 0.1] & 0.7780\% & 0.7986\% & 0.8616\% & 0.6096\% & \textbf{0.5820\%} \\
                      & (0.1, 0.2] & 1.474\%  & 1.557\%  & 1.472\%  & 1.277\%  & \textbf{1.217\%}  \\
                      & (0.2, 0.3] & 2.381\%  & 2.500\%  & 2.295\%  & 2.118\%  & \textbf{2.007\%}  \\
                      & (0.3, 0.4] & 3.276\%  & 3.395\%  & 3.090\%  & 2.917\%  & \textbf{2.763\%}  \\
                      & (0.4, 0.5] & 4.044\%  & 4.185\%  & 3.806\%  & 3.612\%  & \textbf{3.415\%}  \\ \hline
    \multirow{5}{*}{\rotatebox{90}{PSNR(dB)$\uparrow$}} & 
                        (0.0, 0.1] & 29.82    & 29.81    & 29.21    & 30.76    & \textbf{31.16}    \\
                      & (0.1, 0.2] & 26.23    & 25.72    & 26.47    & 26.83    & \textbf{27.14}    \\
                      & (0.2, 0.3] & 23.73    & 23.19    & 24.23    & 24.30    & \textbf{24.66}    \\
                      & (0.3, 0.4] & 22.19    & 21.53    & 22.77    & 22.66    & \textbf{23.03}    \\
                      & (0.4, 0.5] & 21.18    & 20.53    & 21.88    & 21.63    & \textbf{22.01}    \\ \hline
    \multirow{5}{*}{\rotatebox{90}{SSIM $\uparrow$}} & 
                        (0.0, 0.1] & 0.9535   & 0.9591   & 0.9532   & 0.9633   & \textbf{0.9647}   \\
                      & (0.1, 0.2] & 0.9101   & 0.9179   & 0.9158   & 0.9248   & \textbf{0.9282}   \\
                      & (0.2, 0.3] & 0.8563   & 0.8682   & 0.8688   & 0.8783   & \textbf{0.8838}   \\
                      & (0.3, 0.4] & 0.8023   & 0.8204   & 0.8222   & 0.8327   & \textbf{0.8408}   \\
                      & (0.4, 0.5] & 0.7520   & 0.7740   & 0.7766   & 0.7892   & \textbf{0.7999}   \\ \hline
    \multirow{5}{*}{\rotatebox{90}{FID $\downarrow$}} & 
                        (0.0, 0.1] & 19.70    & 14.65    & 18.25    & 14.14    & \textbf{13.47}    \\
                      & (0.1, 0.2] & 31.13    & 27.56    & 26.51    & 26.12    & \textbf{24.57}    \\
                      & (0.2, 0.3] & 49.17    & 42.49    & 41.60    & 41.06    & \textbf{39.41}    \\
                      & (0.3, 0.4] & 64.27    & 56.35    & 54.48    & 54.66    & \textbf{52.65}    \\
                      & (0.4, 0.5] & 80.68    & 71.17    & 69.25    & 68.68    & \textbf{67.25}    \\
                      \hline\hline
    \end{tabular}
    \label{table:quantitative_ratio_places2}
\end{table}

\textbf{Quantitative comparison} For quantitative analysis, we compare our model with current state-of-the-art methods on CelebA-HQ and Places2 datasets. We test the models on masks with random size and masks with certain mask ratios, and evaluated the inpainting results in terms of mean absolute error (MAE), peak signal-to-noise ratio (PSNR), structural similarity index measure (SSIM) and Fréchet inception distance (FID) \cite{NIPS2017_8a1d6947}. We average the performance over 2000 and 36500 images from validation sets of CelebA-HQ and Places2, respectively. The results on CelebA-HQ with randomly generated masks are reported in Table \ref{table:quantitative_celebahq}. It shows that all learning-based models perform better than PatchMatch, and the superiority of the proposed model in the quantitative comparison. In addition, the results in Table \ref{table:quantitative_ratio_places2} show that our model achieves better quantitative performance on Places2 under different mask ratios.

\begin{figure*}[!ht]
    \centering
    \null\hfill
    \subfloat[Input]{\includegraphics[width=0.20\linewidth]{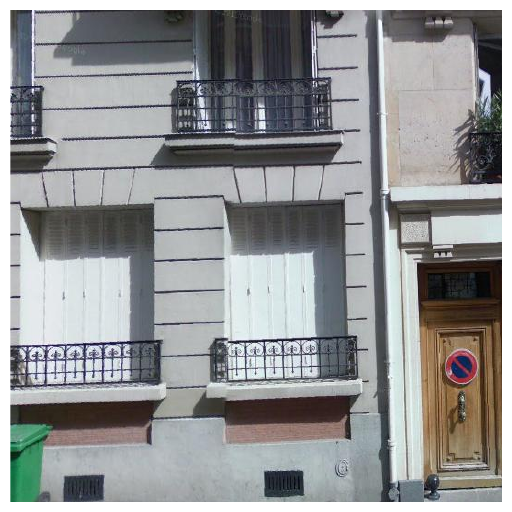}\label{subfig:user_img}}
    \hfill
    \subfloat[Step 1: Edge extraction]{\includegraphics[width=0.20\linewidth]{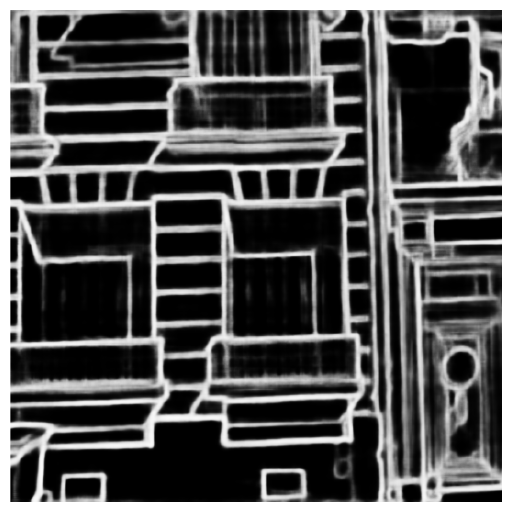}\label{subfig:user_edge}}
    \hfill
    \subfloat[Step 2: Binarization]{\includegraphics[width=0.20\linewidth]{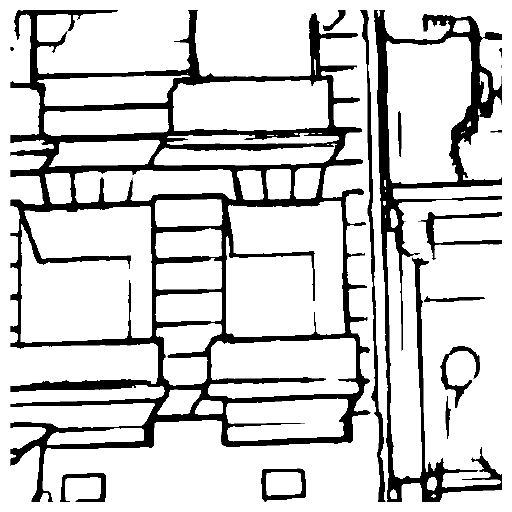}\label{subfig:user_biEdge}}
    \hfill
    \subfloat[Step 3: Skeletonization]{\includegraphics[width=0.20\linewidth]{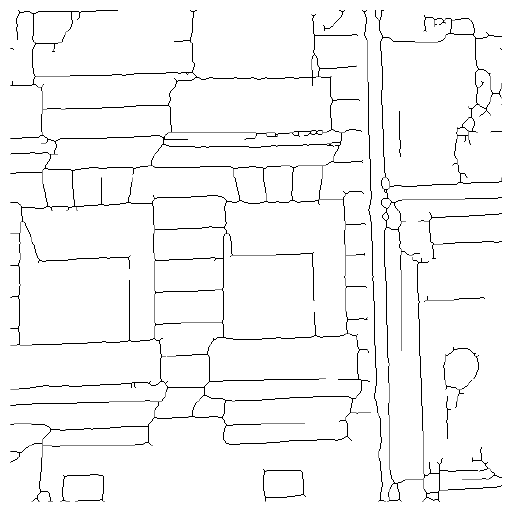}\label{subfig:user_skeleton}}
    \hfill\null
    \caption{Sketch examples generated from the different steps of our sketch mask generation method. Our method uses binalization and skeletonization in step 2 and 3 respectively to refine the edge maps extracted using PiDiNet \cite{Su_2021_ICCV}, and  generates sketches with thiner lines and more clear object structures.}
    \label{fig:sketch}
\end{figure*}

\subsection{User-Guided Inpainting}
To extend the proposed network for user-guided image inpainting, which provides interactive scenes to allow users to control the inpainting results, we adopt sketch generation and use the sketch-based binary mask as an additional input during training.  Our sketch mask generation method consists of three steps: 1) generating gray edge maps; 2) converting the gray maps to binary images; 3) improving the binary images with clearer lines, e.g., thinner with less adhesion. We directly use the edge detector PiDiNet in \cite{Su_2021_ICCV} to generate a gray edge map, and produce a binary image by setting all values in the gray map above a constant threshold (e.g., 0.65 for an image from Paris StreetView dataset) to ones and the others to zeros. Next, we perform area opening \cite{10.1007/978-3-662-03039-4_13} on the binary image to remove small objects and generate a new binary image, i.e., removing all connected components/objects that have fewer than 100 pixels in 8-connectivity. Finally, we perform morphological skeletonization \cite{1164959} to further extract skeleton of figures in the binary image. Sketch examples generated from the different steps are shown in Fig. \ref{fig:sketch}.

\begin{figure}[!ht]
    \centering
    \setlength\tabcolsep{1pt}
    \captionsetup{justification=centering}
    \begin{tabular}{cccc}
        \subfloat[Input w/o guide]{\includegraphics[width=0.24\linewidth]{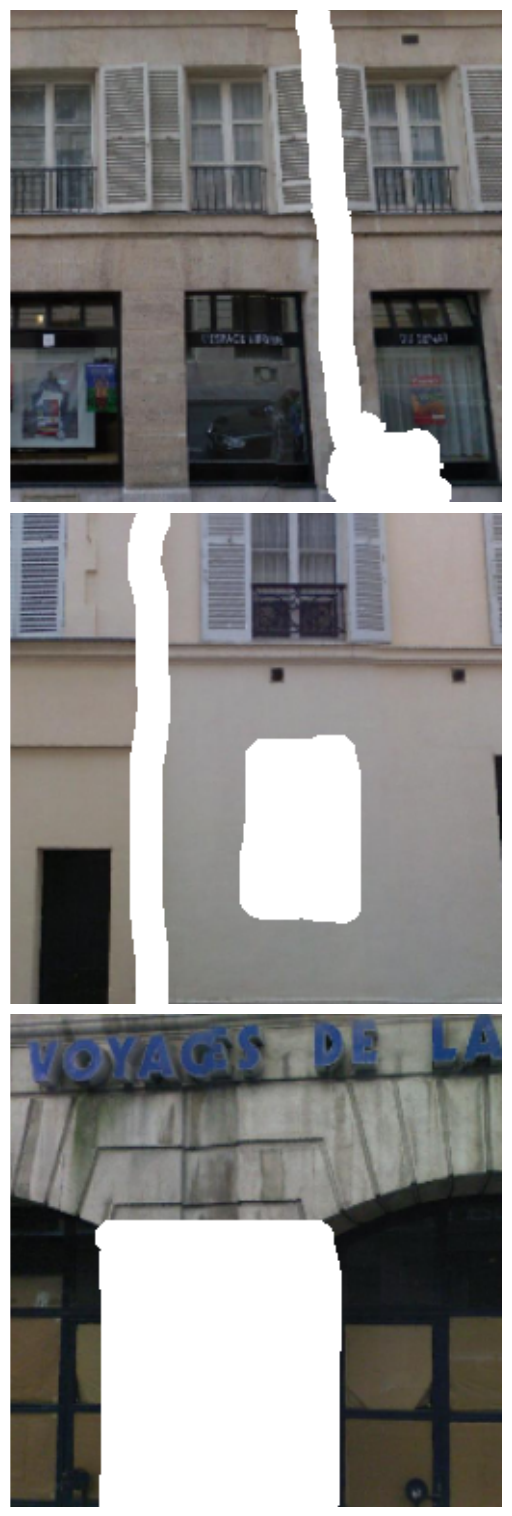}\label{subfig:sketch_in_wo}}
        \subfloat[Result w/o guide]{\includegraphics[width=0.24\linewidth]{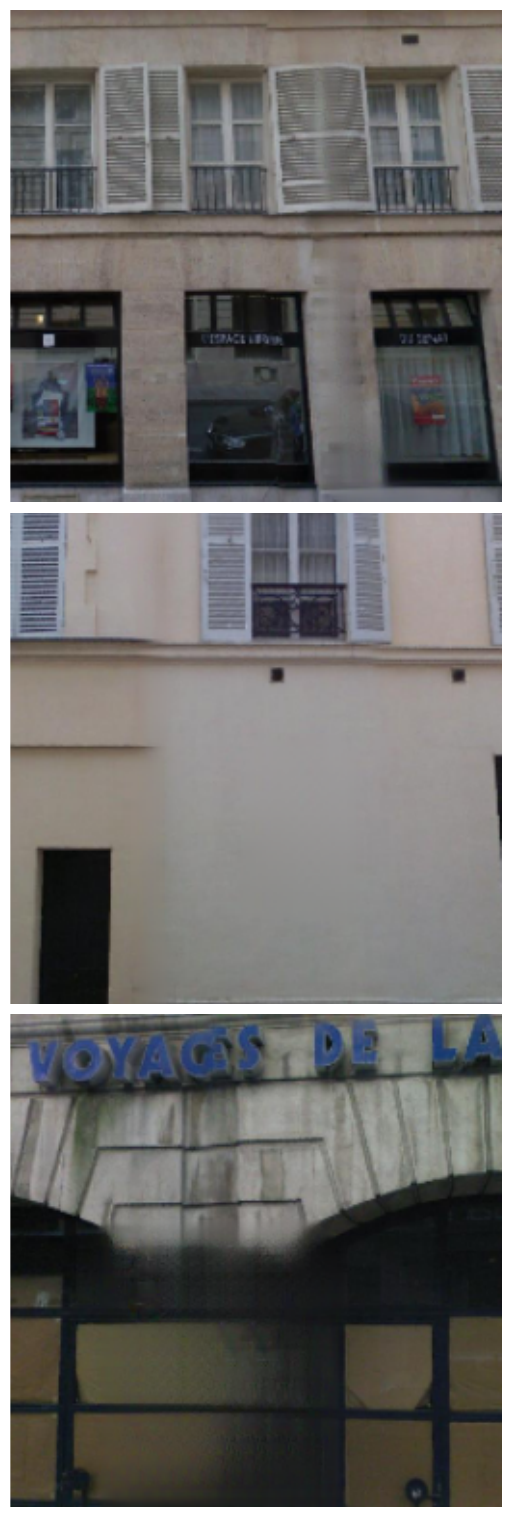}\label{subfig:sketch_out_wo}}
        & 
        \tikz{\draw[-, black, dash pattern=on 5pt off 5pt, thick](0,1.5) -- (0,7.55);}
        &
        \subfloat[Input w/ guide]{\includegraphics[width=0.24\linewidth]{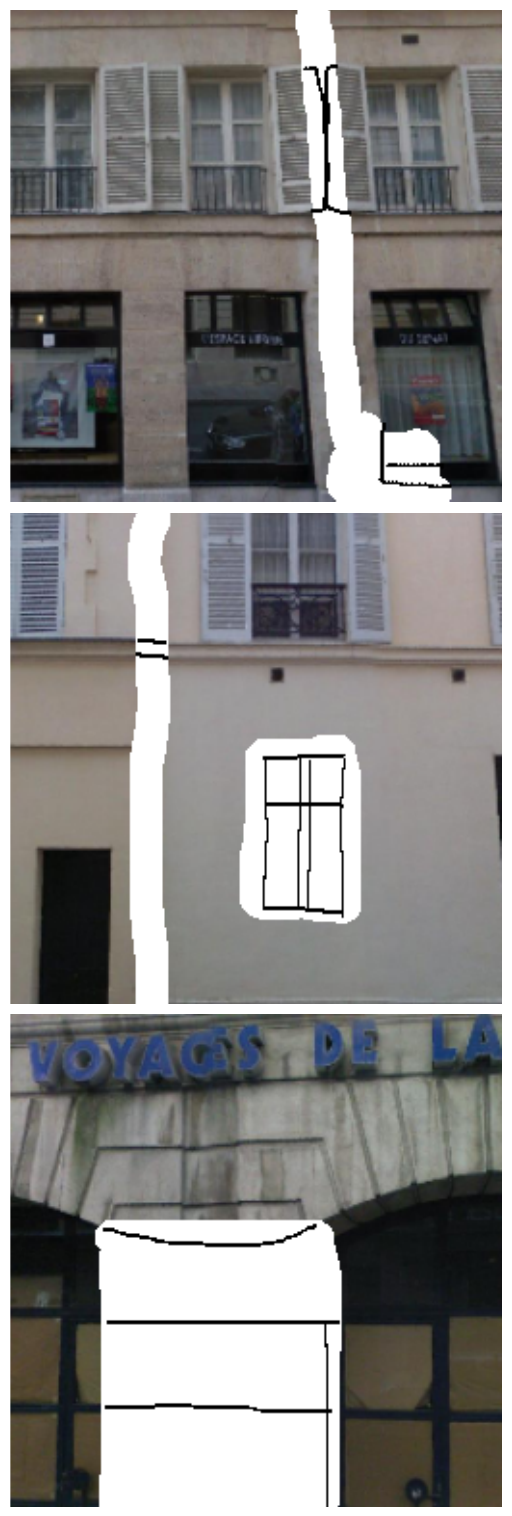}\label{subfig:sketch_in_w}}
        \subfloat[Result w/ guide]{\includegraphics[width=0.24\linewidth]{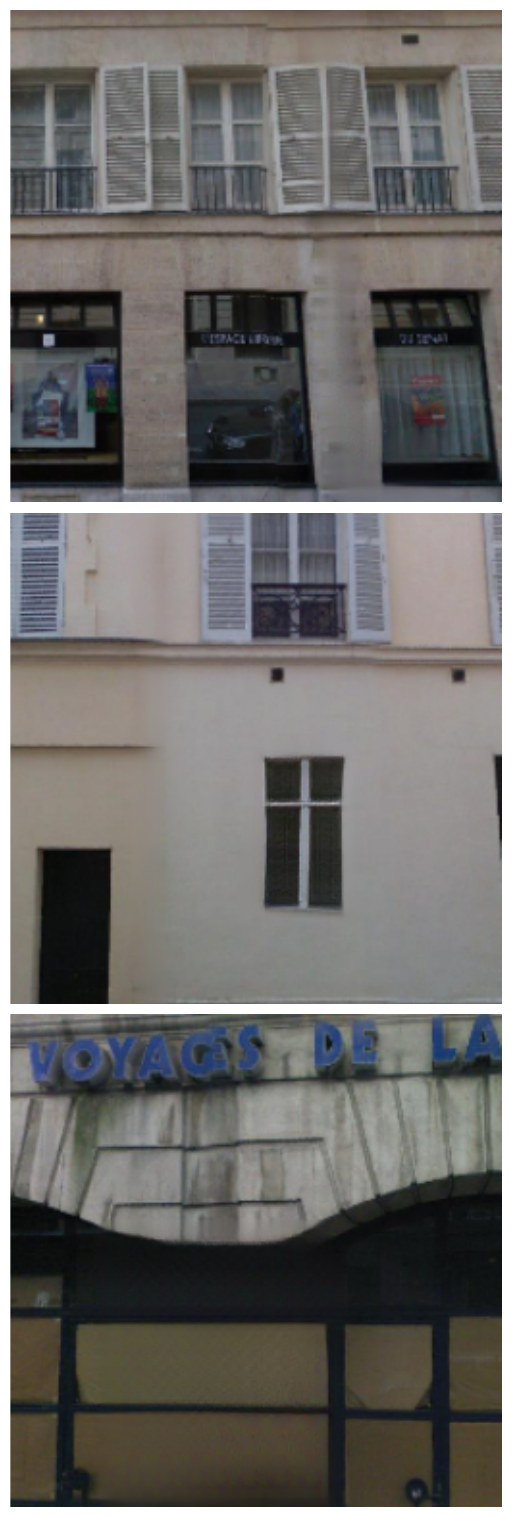}\label{subfig:sketch_out_w}}
    \end{tabular}
    \caption{Results of our user-guided inpainting. Our model can generate more desired inpainting results according to the user-provided sketches.}
    \label{fig:guide}
\end{figure}

To train our user-guided inpainting model, we generate a sketch-based binary mask by multiplying the sketch mask with the binary mask $\mathbf{M}$, and use it as an additional channel input. The network architecture of the user-guided inpainting model is the same as those of the proposed model, but trained with 5-channel inputs, including R, G, B channels, binary mask channel and sketch-based binary mask channel. During testing, users draw lines as sketch or edge to guide the inpainting model to generate the desired inpainting results. The examples of user-guided inpainting results on Paris StreetView in Fig. \ref{fig:guide} show that our user-guided inpainting model can generate visually appealing contents according to the user guidance.

\section{Conclusion}
In this work, we proposed an enhanced free-form image inpainting framework by identifying and addressing three issues in our previous work \cite{zhou2021atma}, i.e., temperature training stability, computational efficiency of the attention module, and visual artifacts. For stabilizing temperature training, we designed a temperature learning network with Softplus activation function. The new temperature learning network avoids to generate negative temperatures that results in unstable training, and it is able to learn multiple adaptive temperatures that help the inpainting framework capture more long-range contextual information and generate high quality inpainting results. To improve computational efficiency of the attention module with larger patch size and parallel computing scheme, a masked self-attention mechanism that formulates features as queries, keys and values and does dot-product attention was introduced. In addition, we further improved realism and appearance consistency of the generated results by modifying the network architecture, e.g., removing batch normalization, and using Hinge Adversarial Loss with discriminator Spectral Normalization (HAL-SN). 

 The proposed image inpainting framework was analyzed and evaluated on three benchmark image inpainting datasets, including Paris StreetView, CelebA-HQ and Places2. Experimental results demonstrated the effectiveness of the proposed framework on temperature learning and image inpainting. Furthermore,  experiments on quantitative and qualitative comparisons between the proposed method and the state-of-the-art free-form image inpainting methods in referenced literature demonstrated the superiority of the proposed method. Moreover, we presented a sketch generation method and extended our inpainting framework for user-guided image inpainting. Our experimental results on Paris StreetView showed that the proposed framework with user sketches as an additional input allows users to have fine control of the final results. 

The proposed framework can generate high quality inpainting results and can be used for user-guided image inpainting, which has an important value for image inpainting and interactive image editing. However, similar to conventional nearest neighbor based patch matching methods, doing pixel-wise nearest neighbor based patch-matching in deep image generative models for high quality image generation is not memory efficient. Feature work will be interesting to prune the network architecture to improve memory efficiency of the patch-based attention algorithm. Furthermore, considering the practical deployment of our inpainting model, we feel that it will be important to find new ways to reduce the training data requirements in our future studies.

\bibliographystyle{ieeetr}
\bibliography{reference}

\clearpage

\end{document}